%% file: main.tex
\documentclass{article}

\usepackage{format/iclr2027_conference}
\usepackage{times}

\usepackage[T1]{fontenc}
\usepackage{microtype}

\usepackage{amsmath}
\usepackage{amssymb}
\input{format/math_commands.tex}   
\def\vdelta{{\bm{\delta}}}          

\usepackage{graphicx}
\usepackage{booktabs}
\usepackage{longtable}
\usepackage{array}
\usepackage{flafter}
\usepackage{placeins}
\usepackage{algorithm}
\usepackage{algpseudocode}
\algrenewcommand{\algorithmicrequire}{\textbf{Input:}}
\algrenewcommand{\algorithmicensure}{\textbf{Output:}}

\usepackage{xcolor}
\usepackage{url}
\usepackage{hyperref}

\definecolor{writepink}{RGB}{255,20,147}

\iclrfinalcopy

\title{Test-Time Compute for Tabular Foundation Models: Mechanisms, Gains, and Limits}

\author{%
\begin{tabular}{@{}c@{}}
\makebox[\dimexpr\textwidth-2\tabcolsep\relax][c]{%
\begin{tabular}{@{}cccc@{}}
Kanghui Ning\textsuperscript{1} &
Marin Bilo\v{s}\textsuperscript{2} &
James T. Wilson\textsuperscript{2} &
Yilang Zhang\textsuperscript{2}
\end{tabular}
}\\[0.35em]
\makebox[\dimexpr\textwidth-2\tabcolsep\relax][c]{%
\begin{tabular}{@{}cccc@{}}
Kashif Rasul\textsuperscript{2} &
Dongjin Song\textsuperscript{1} &
Anderson Schneider\textsuperscript{2} &
Yuriy Nevmyvaka\textsuperscript{2}
\end{tabular}
}\\[0.6em]
\makebox[\dimexpr\textwidth-2\tabcolsep\relax][c]{%
\normalfont\small
\begin{tabular}{@{}c@{}}
\textsuperscript{1}School of Computing, University of Connecticut, Storrs, USA.\\
\textsuperscript{2}Department of Machine Learning Research, Morgan Stanley, New York, USA.
\end{tabular}
}
\end{tabular}
}

\begin{document}

\maketitle
\lhead{}

\begin{abstract}
Which forms of test-time compute improve the predictions of strong pretrained tabular foundation models (TFMs)? We systematically study this along three axes: adaptation, aggregation, and context construction. Our evaluation spans modern TFMs across the TabArena benchmark, supplemented by experiments on wide and large-scale tables from OpenML.
For adaptation, we introduce DiagScale, a diagonal query--key similarity update.
It trains only 0.003--0.03\% of model parameters and achieves gains comparable to full fine-tuning across three independently pretrained backbones.
For aggregation, both pool composition and selection strategy matter.
TabPFN-3 already averages predictions from different preprocessing variants of the same data, and adding more such predictions yields diminishing returns.
With a broader pool of 96 configurations, greedy selection reduces error by 2.4\% relative to the default predictor, but uniform averaging increases error.
For context construction, attention-guided retrieval improves TabPFN-3's predictions on some large tables and supports source pools beyond the full context memory limit.
The context expansion methods we test yield no consistent improvement.
Taken together, our results suggest that adaptation and selective aggregation yield consistent benchmark-level gains.
The benefits of context construction depend more on the task and data regime.
Adaptation and aggregation over the same backbone yield further gains when combined, but require substantially more computation than default inference.
These trade-offs motivate choosing strategies according to the available computation budget.
Code is available at \url{https://github.com/kanghui-learning/test-time-compute-for-tabular-foundation-models}.
\end{abstract}

\vspace{-4mm}
\input{introduction}

\input{related_work}

\input{framework}

\input{evaluation}

\input{adaptation}

\input{aggregation}

\input{context_construction}

\input{analysis}

\input{discussion}


\clearpage
\section*{AI use statement}
In this work, we used generative AI tools to assist with drafting and polishing parts of the text, retrieving and discovering related work, and writing and refining parts of the software code. We have reviewed all AI-assisted work. We take responsibility for the final content of this work, including text, claims or artifacts produced with the aid of generative AI.

\section*{Reproducibility statement}
The main text and appendices describe the model configurations, dataset splits, adaptation and aggregation procedures, and statistical analyses used in our experiments. Additional implementation details and results are provided in Appendices A–E, including runtime measurements and controlled comparisons.

\bibliography{main}
\bibliographystyle{format/iclr2027_conference}

\clearpage
\appendix
\AddToHook{env/table/begin}{\setlength{\belowcaptionskip}{4pt}}

\input{appendix_evaluation}

\input{appendix_adaptation}

\input{appendix_aggregation}

\input{appendix_context}

\input{appendix_cross_analysis}

\end{document}

%% file: format/math_commands.tex
\usepackage{amsmath,amsfonts,bm}

\def\eqref#1{equation~\ref{#1}}

\def\1{\bm{1}}

\def\vzero{{\bm{0}}}

\def\vb{{\bm{b}}}

\def\vg{{\bm{g}}}

\def\vk{{\bm{k}}}

\def\vq{{\bm{q}}}

\def\vs{{\bm{s}}}

\def\vw{{\bm{w}}}
\def\vx{{\bm{x}}}

\def\mI{{\bm{I}}}

\DeclareMathAlphabet{\mathsfit}{\encodingdefault}{\sfdefault}{m}{sl}
\SetMathAlphabet{\mathsfit}{bold}{\encodingdefault}{\sfdefault}{bx}{n}

\DeclareMathOperator*{\argmin}{arg\,min}

%% file: introduction.tex
\section{Introduction}
\label{sec:intro}

Tabular foundation models (TFMs) have become competitive alternatives to task-specific learners for tabular classification and regression \citep{hollmann2025tabpfn, erickson2025tabarena}. Recent studies have explored ways to improve predictions through task-specific training and inference. For example, \citet{rubachev2025finetuning} compare fine-tuning strategies for TabPFN-v2 and analyze how
  adaptation changes its attention, while LoCalPFN combines local retrieval with fine-tuning \citep{thomas2024localpfn}. In such a combined pipeline, gains can come from the selected context, the
  parameter updates, or the interaction between them. This motivates examining the contribution of individual operations alongside the performance of the complete method.

The capabilities of the underlying models have also changed substantially. The original TabPFN targeted datasets with up to 1,000 rows and 100 features \citep{hollmann2023tabpfn}, whereas
  TabPFN-3 reports strong performance at 1M rows with 200 features and averages eight predictions by default \citep{priorlabs2026tabpfn3}. 
  As tabular foundation models grow increasingly capable out of the box, the extent to which additional test-time computation improves their predictions remains poorly understood.
  We therefore systematically study additional task-specific computation on
  frontier TFMs through controlled comparisons, examining which design choices improve predictions and what limits the gains.

We use \emph{test-time computation} (TTC) to mean computation allocated to a deployed task after
pretraining, including task-specific fitting and additional inference. We factorize the prediction
to distinguish three modifiable components: parameters (\emph{adaptation}), member predictions
and their reduction (\emph{aggregation}), and conditioning data (\emph{context construction}).
This framework guides controlled comparisons of what individual interventions contribute and
whether combining them provides further gains. We study adaptation across TabPFN-3, TabICL v2,
and TabFM, and examine aggregation and context construction in depth on TabPFN-3. Our evaluation
combines TabArena with additional wide and large tables from OpenML. The main findings are as follows.

\vspace{-3mm}
\paragraph{Compact similarity adaptation is effective across backbones.}
A 25-setting adaptation sweep and targeted component experiments motivate restricting the update
to query--key similarity. We develop \texttt{DiagScale}, which learns a diagonal multiplier in each attention
module while freezing pretrained weights. It trains $0.003$--$0.03\%$ of the parameters and achieves
gains close to full fine-tuning on all three backbones. These results suggest that a small
set of learned similarity corrections can capture much of the adaptation gain on this benchmark.

\vspace{-3mm}
\paragraph{Aggregation benefits from both diversity and selection}
Increasing TabPFN-3's native ensemble gives diminishing returns. Greedy selection from a pool of
96 configurations gains $67$ Elo, whereas uniform averaging over the same predictions degrades predictive performance. The larger pool is useful when paired with a reducer that selects among its members,
but restricting the selector's access to training-side labels substantially reduces this advantage.

\vspace{-3mm}
\paragraph{Changing the context can help or hurt predictions.}
Attention-guided row selection can improve predictions when the full training set fits within
the context capacity and support retrieval from larger pools. In matched comparisons, attention-guided row selection can improve or degrade performance relative to random selection. These contrasting results suggest that the benefits of context selection depend on
how the selection rule interacts with data characteristics. The benefits of column selection
also vary across datasets, while the tested expansion methods, including synthetic rows,
pseudo-labeled examples, and derived features, provide no consistent improvement over the
original context.

\vspace{-3mm}
\paragraph{Combining interventions provides further gains.}
Combining parameter-efficient fine-tuning (\texttt{DiagScale}) with configuration aggregation reaches
$+93$ Elo on TabPFN-3, improving over either intervention alone. On the same benchmark workload,
the combination costs 253 GPU-hours, compared with 6.6 for DiagScale and 248 for aggregation.
Adaptation spends most of its compute on task-specific fitting, whereas the evaluated aggregation
pipeline generates all 96 configurations' predictions for selection and evaluation. The additional
gain therefore comes with substantially more measured computation than adaptation alone.

%% file: related_work.tex
\vspace{-3mm}
\section{Related work}
\label{sec:related}

\vspace{-3mm}
\paragraph{Tabular foundation models.}
Prior-data fitted networks amortize supervised learning through pretraining on synthetic tasks
\citep{muller2022bayesian}. TabPFN and its successors, including TabPFN-3.5, have made this
approach competitive with strong tabular learners
\citep{hollmann2023tabpfn, hollmann2025tabpfn, priorlabs2026tabpfn3, jaeger2026tabpfn35}.
Independently pretrained models include TabICL, TabFM, and LimiX-2
\citep{qu2025tabicl, qu2026tabiclv2, google2026tabfm, zhang2026limix2}; other lines extend transfer across tables,
context capacity, and semantic information \citep{ma2024tabdpt, zhu2023xtab, kim2024carte,
arazi2025tabstar, spinaci2025contexttab}. TabArena standardizes comparisons on curated real-world
tasks and reusable predictions \citep{erickson2025tabarena}, building on TabRepo's evaluation
infrastructure \citep{salinas2023tabrepo}.

\vspace{-3mm}
\paragraph{Task-specific learning and inference.}
Prior work improves pretrained tabular foundation models through task-specific optimization and inference
procedures. \citet{rubachev2025finetuning} compare fine-tuning strategies for TabPFN-v2 and
analyze how fine-tuning changes the attention-based weighting of in-context examples.
Context optimization and retrieval address limitations in context capacity and the choice
of conditioning examples. TuneTables learns a compact context \citep{feuer2024tunetables},
while LoCalPFN and MixturePFN combine context selection with task-specific parameter updates
\citep{thomas2024localpfn, xu2024mixturepfn}. LimiX uses attention-guided retrieval within its
inference pipeline \citep{zhang2025limix}.
VIP-COP selects training examples and features based on their estimated importance for prediction
\citep{chen2026vipcop}.
DistPFN adjusts predicted class probabilities at test time to mitigate label shift
\citep{lee2026mitigating}.

Ensembling provides another way to improve predictions through additional computation
\citep{erickson2020autogluon, ning2026timerouter}.
TabPFN averages predictions across preprocessing choices and permutations
\citep{hollmann2025tabpfn, ye2025closerlook}, while greedy ensemble selection combines
selected members from \mbox{a model library \citep{caruana2004ensemble}}. Tuning and ensembling
are established components of AutoGluon and TabArena
\citep{erickson2020autogluon, erickson2025tabarena}.
These practices motivate examining both how candidate predictions are generated and how
they are selected or combined.

Against this background, we systematically investigate the benefits and limitations of
additional task-specific computation on strong pretrained tabular foundation models.
Section~\ref{sec:framework} organizes this investigation around the components of the inference pipeline.

\vspace{-3mm}
\paragraph{Test-time compute.}
Research on language-model test-time compute studies additional sampling, search, and
verification \citep{wang2023selfconsistency, snell2024scaling, wu2024inference}.
Test-time training provides a related direction in which model parameters are updated using
deployment data \citep{sun2020ttt, wang2021tent}.
Our study takes a task-level view of post-pretraining computation, covering both task-specific adaptation on labeled training data and inference on new/unseen queries.

%% file: framework.tex
\vspace{-3mm}
\section{A factorized view of test-time compute}
\label{sec:framework}
\label{sec:factorization}
\begin{figure}[htbp]
\centering
\includegraphics[width=\linewidth]{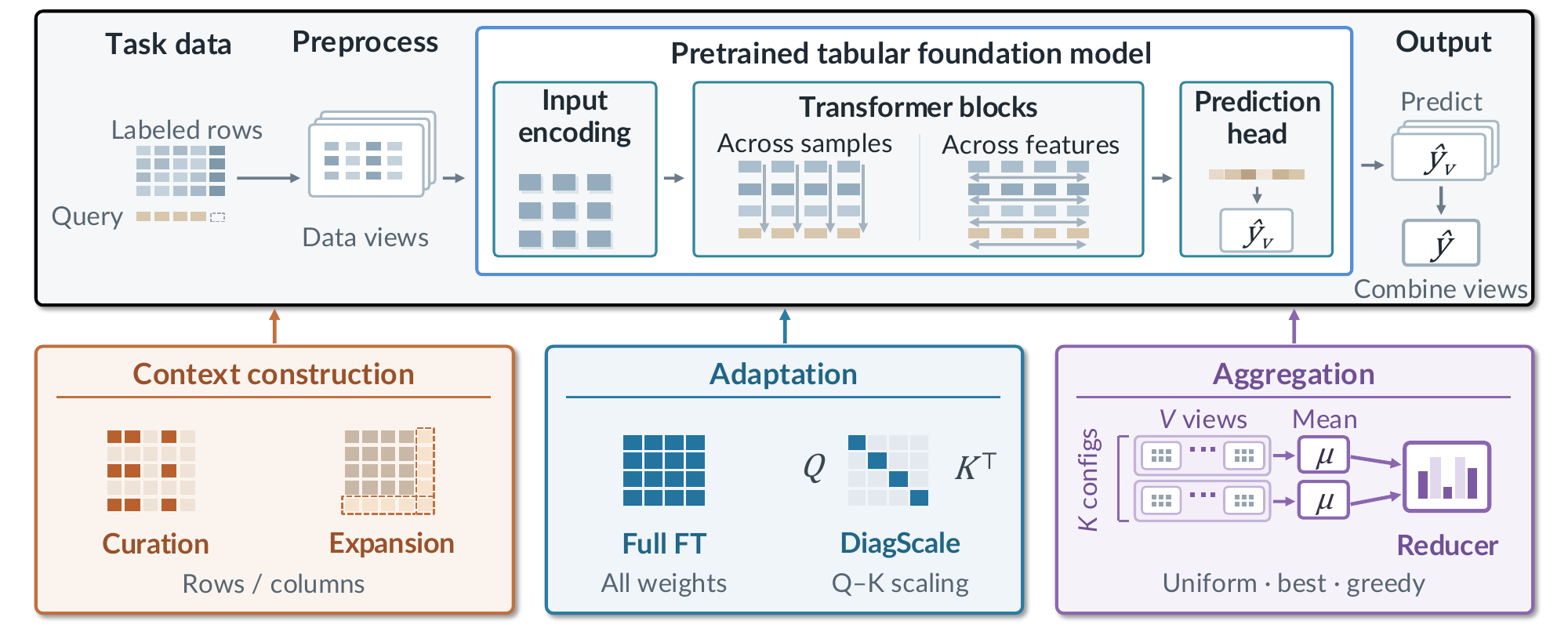}
\caption{Three families of test-time compute modify different components of the inference pipeline. Top: an inference pipeline comprising preprocessing, a frozen
tabular foundation model, and combination of predictions across data views.
Bottom: context construction changes the conditioning data; adaptation changes
model parameters; aggregation changes the prediction pool and the rule used to
select or combine its members.}
\label{fig:framework-overview}
\end{figure}

We organize TTC around three aspects of the inference pipeline: the model parameters, the
collection and combination of member predictions, and the conditioning data. These correspond to
adaptation, aggregation, and context construction, respectively
(Figure~\ref{fig:framework-overview}).

For labeled training data $D$ and a query $\vx$, we express the inference pipeline as
\begin{equation}
\label{eq:anatomy}
\hat y(\vx)=\mathcal R_{\vw}\!\left(\left\{
 f_{\theta_m}\big(\vx;g_m(D,\vx)\big)\right\}_{m=1}^{M}\right).
\end{equation}
Here $\theta_m$ is a pretrained or adapted model state; $g_m$ constructs the conditioning rows,
columns, and representation; and $\mathcal R_{\vw}$ combines the $M$ member predictions.
The labeled training data may be partitioned for task-specific fitting and validation.
Adapted states and fitted reducer weights depend on $D$, and a constructor that selects rows using
the model's attention also depends on $\theta_m$ (Section~\ref{sec:context}). Methods may modify
several components.

\label{sec:implications}
\textbf{Adaptation} changes $\theta_m$. Restricting the updated parameters controls the family of
task-specific corrections and the amount of information needed to estimate them. We compare
parameter subsets and test whether a similarity-only correction is sufficient for useful gains.

\textbf{Aggregation} changes the members and their reduction. Individual quality and error
complementarity determine what a pool can offer; the reducer and its validation data determine
which combination is selected. Uniform averaging uses fixed weights, whereas selecting members
or learning their weights requires validation data. This introduces additional design choices
concerning the reduction rule and the use of validation data.

\textbf{Context construction} changes $g_m$. We investigate two directions. First, curation and
retrieval select existing rows or columns to form a support set for in-context learning,
emphasizing relevant evidence or satisfying capacity constraints. Second, context expansion
derives synthetic rows, columns, or labels; we test whether adding this context improves the
in-context learning performance of TFMs.

%% file: evaluation.tex
\vspace{-3mm}
\section{Evaluation design}
\label{sec:setup}

\vspace{-3mm}
We study adaptation on TabPFN-3, TabICL v2, and TabFM
\citep{priorlabs2026tabpfn3, qu2026tabiclv2, google2026tabfm}, and aggregation and context
construction on TabPFN-3. We mainly use TabArena \citep{erickson2025tabarena} as our evaluation
benchmark, which comprises 51 classification and regression datasets ranging from 748 to
150{,}000 rows and covering ten domains, including business and marketing, finance, healthcare,
and the natural sciences.
We follow its official evaluation protocol (Appendix~\ref{app:datasets}).
Additional datasets for our wide-table and large-row studies are collected from OpenML and
evaluated using the splits and reference models specified in Appendices~\ref{app:datasets}
and~\ref{app:context}.

We fit Elo ratings from wins, ties, and losses between methods evaluated on the same
test splits; Elo rating differences model their expected probabilities of winning
against each other \citep{erickson2025tabarena}.
We evaluate candidates against 68 fixed reference entries
(Table~\ref{tab:reference-models}), with the fitting setup specified
in each experiment. $\Delta$ Elo compares the candidate with
its baseline within the same fit.
We also report TabArena's normalized score from the same comparison set, which
rescales performance on each dataset to $[0,1]$ before averaging,
and percentage changes in error relative to the baseline.
Appendices~\ref{app:evidence-tiers} and~\ref{app:evaluation-grid} detail the metrics,
aggregation procedures, and evaluation coverage.

%% file: adaptation.tex
\vspace{-3mm}
\section{Adaptation: a compact similarity update}
\label{sec:adaptation}
\vspace{-3mm}
\paragraph{Which parameters should we update?}
\citet{rubachev2025finetuning} identify full fine-tuning as a practical baseline on TabPFN-v2.
We revisit and extend their adaptation menu on TabPFN-3, comparing 25 settings ranging from
1,280 trainable parameters to full fine-tuning. These include LoRA, bias-only tuning, and updates
restricted to LayerNorms, embeddings, or selected transformer blocks.
Table~\ref{tab:full-ladder} lists all settings and their results. The sweep asks which parts of
the model need to change to obtain the gains of adaptation.

We use a single learning rate for each backbone--adaptation pair across datasets, following
the idea of shared hyperparameter defaults \citep{probst2019tunability, holzmuller2024better}.
Learning rate grids and selection criteria are given in Appendix~\ref{app:adaptation-sweeps}.
Within each run, we select the checkpoint using inner validation and fall back to the pretrained
weights if no checkpoint improves validation performance over the pretrained model
(Appendix~\ref{app:adaptation-protocol}).
We evaluate each recipe by adding it separately to the same
68 reference entries (Table~\ref{tab:reference-models}). Each $\Delta$ Elo is measured against
the reference set's published TabPFN-3 default within that recipe's fit.

SoftScale stands out in this menu (Table~\ref{tab:peft-ladder}). Updating only the pretrained
query-scaling modules reaches $+26$ Elo with 1.03M parameters, close to full fine-tuning
at $+25$ with 58.3M; their normalized scores are $0.703$ and $0.704$, respectively. This motivates us to examine
what the query-side scaling changes and whether a simpler update can retain its benefit.

\vspace{-3mm}
\begin{table}[!htbp]
\centering
\small
\caption{Selected adaptation recipes on TabPFN-3. LoRA uses the rank with the highest Elo.
Each fit contains one recipe and the same 68 references;
$\Delta$ Elo is relative to the published TabPFN-3 default in that fit.
Parameter counts use the regression checkpoint.}
\label{tab:peft-ladder}
\begin{tabular}{@{}llrrr@{}}
\toprule
\textbf{Setting} & \textbf{Updated} & \textbf{Params} & \textbf{$\Delta$ Elo} & \textbf{Norm. score} \\
\midrule
Full fine-tuning & all trainable weights & 58.3M & $+25$ & $\mathbf{0.704}$ \\
LoRA ($r=16$)   & low-rank updates      & 3.51M & $+15$ & $0.684$ \\
Value--output    & $W_V$, $W_O$          & 12.9M & $+19$ & $0.693$ \\
Query--key       & $W_Q$, $W_K$          & 12.9M & $+22$ & $0.696$ \\
\textbf{SoftScale} & query-side scaling  & 1.03M & $\mathbf{+26}$ & $0.703$ \\
\bottomrule
\end{tabular}
\end{table}

\begin{figure}[!t]
\centering
\includegraphics[width=\linewidth,trim=0 4bp 0 0,clip]{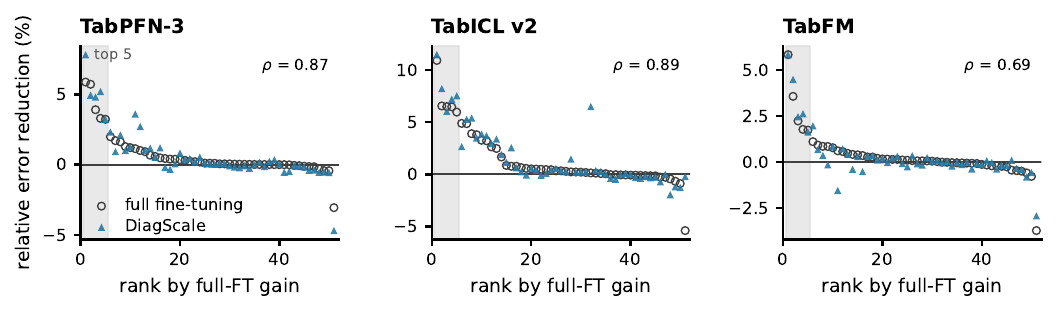}
\caption{Per-dataset error reduction for DiagScale and full fine-tuning, relative to the same
frozen backbone. Each panel orders 51 datasets by full fine-tuning's reduction; shading marks its
top five. $\rho$ is the Spearman correlation between the two methods' error reductions.}
\label{fig:diagscale-route}
\end{figure}

\vspace{-4mm}
\paragraph{What SoftScale fine-tuning changes.}
To improve generalization to longer contexts, TabICL v2 introduced query-aware
scalable softmax, which counteracts attention fading as context length grows
\citep{qu2026tabiclv2}. TabPFN-3 adopts this mechanism
\citep{priorlabs2026tabpfn3}; we refer to fine-tuning its pretrained scaling
modules as SoftScale. Within attention layers, each scaling module transforms a query as
$\tilde{\vq}=\vq\odot\vs(n)\odot\vg(\vq)$, where $n$ is the number of keys in the attention
context, $\vs(n)$ is a learned length-dependent scale, and $\vg$ is a query-dependent gate.
On three diagnostic datasets, we separately fine-tune the networks producing $\vs$ and $\vg$,
keeping all other parameters frozen. Training the context-size branch $\vs$ alone recovers
$93$--$106\%$ of the gain from jointly fine-tuning both branches, compared with $8$--$16\%$
for the query-dependent gate $\vg$ (Appendix~\ref{app:adaptation-mechanisms},
Table~\ref{tab:softscale-branch-training}).

We further examine how fine-tuning changes the scale vectors on these diagnostic datasets.
The learned changes primarily alter the relative scaling of individual dimensions, rather than
uniformly rescaling each vector (Appendix Figure~\ref{fig:diagscale-mech}B).
Together with the preceding results, these observations motivate learning dimension-specific
query multipliers directly, providing an explicit way to adapt query--key similarity that can
also be applied to other TFMs.

\vspace{-3mm}
\paragraph{DiagScale.}
For a fixed context size, the scale branch produces one vector per module. Alternatively, we can learn a correction
of this form directly, freezing every pretrained parameter. With $\vq$ denoting the query after
its pretrained scaling and $\vk$ the key, the modified attention logit is
\begin{equation}
\label{eq:diagscale}
A_{\vdelta}(\vq,\vk)=\vq^\top\big(\mI+\mathrm{diag}(\vdelta)\big)\vk.
\end{equation}
Each attention module receives a zero-initialized vector $\vdelta$ with one entry per head and
dimension. The initial mapping equals the frozen model; task loss trains only these multipliers,
with training-side checkpoint selection and fallback as specified in
Appendix~\ref{app:adaptation-protocol}.
TabPFN-3 uses 12,672 parameters for regression and 13,056 for classification, including its decoder.
At the attention-logit level, DiagScale is equivalent to the key-scaling component of
IA$^3$ \citep{liu2022ia3}. We investigate whether adapting query--key similarity alone can
provide effective adaptation across TFMs.

\vspace{-4mm}
\begin{table}[!htbp]
\centering
\small
\caption{DiagScale and full fine-tuning across three backbones on 51 datasets. Each Elo fit
contains one adapted model and 68 fixed references (67 TabArena entries plus frozen TabFM);
$\Delta$ Elo is relative to the corresponding reference baseline within that fit.
Parameter counts and percentages use regression checkpoints.}
\label{tab:cross-backbone}
\begin{tabular}{@{}lrrrr@{}}
\toprule
\textbf{Backbone} & \shortstack{\textbf{Backbone}\\\textbf{params}} & \shortstack{\textbf{Trainable params}\\\textbf{(\% of backbone)}} & \shortstack{\textbf{$\Delta$ Elo}\\\textbf{DiagScale}} & \shortstack{\textbf{$\Delta$ Elo}\\\textbf{Full FT}} \\
\midrule
TabPFN-3   & 58.3M & 12.7K ($0.022\%$) & $+26$ & $+25$ \\
TabICL v2  & 28.5M & 7.3K ($0.026\%$) & $+82$ & $+77$ \\
TabFM      & 1.65B & 53.8K ($0.0033\%$) & $+20$ & $+28$ \\
\bottomrule
\end{tabular}
\end{table}

\vspace{-3mm}
\paragraph{Cross-backbone results.}
DiagScale achieves gains close to full fine-tuning across all three backbones
(Table~\ref{tab:cross-backbone}). Its parameter fraction ranges from $0.003\%$ to
$0.03\%$. The per-dataset reductions are correlated with those of full fine-tuning
($\rho=0.69$--$0.89$; Figure~\ref{fig:diagscale-route}). For both methods, larger improvements
are concentrated in a subset of datasets (Figure~\ref{fig:diagscale-route};
Appendix~\ref{app:gain-concentration}).
Direct comparisons with full fine-tuning are reported in Appendix Table~\ref{tab:diagscale-paired}.
Together, these results support adapting query--key similarity as an effective alternative to full
fine-tuning across these TFMs.

%% file: aggregation.tex
\vspace{-2mm}
\section{Aggregation: configuration breadth and selective reduction}
\label{sec:aggregation}

\begin{figure}[!t]
\centering
\includegraphics[width=0.9\linewidth,trim=0 6bp 0 0,clip]{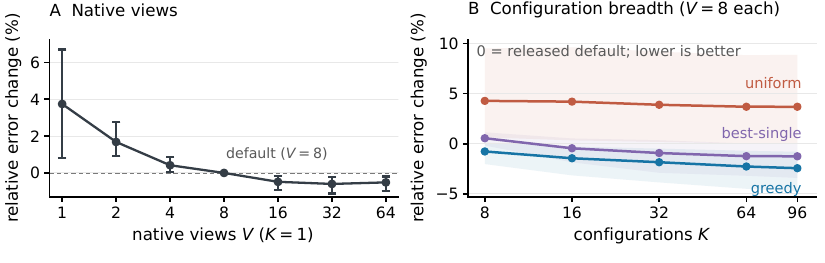}
\vspace{-3mm}
\caption{Aggregation on TabPFN-3. \textbf{A}: varying native views within one configuration.
\textbf{B}: varying configuration count under three reducers, with eight native views per
configuration. Errors are relative to the released eight-view default; negative values indicate
improvement. Error bars and shaded bands show 95\% confidence intervals. Complete curves appear
in Figure~\ref{fig:agg-full}.}
\vspace{-3mm}
\label{fig:agg-canonical}
\end{figure}

\vspace{-2mm}
\paragraph{Two levels of aggregation.}
TabPFN-3 already averages multiple predictions under its default inference settings
\citep{hollmann2025tabpfn, ye2025closerlook, priorlabs2026tabpfn3}.
We investigate how much further aggregation can improve this predictor by increasing the number
of predictions within a configuration or combining predictions across configurations. We also
examine how the resulting gains depend on the rule used to select or combine these predictions.

A \emph{configuration} specifies the inference and preprocessing settings. Under these settings,
the model's built-in ensemble generates $V$ \emph{native views} ($V=8$ by default): predictions obtained with
perturbations such as feature permutations and preprocessing variations. The native aggregation
rule $\mathcal{A}_c$ combines these views into one member prediction $p_c$, using averaging by default.
Across $K$ configurations, a \emph{reducer}
selects or combines these members. Writing $f_{c,v}$ for view $v$ of configuration $c$, and
$P_{\mathcal{I}}=\{p_c:c\in\mathcal{I}\}$ for a pool indexed by $K=|\mathcal{I}|$ configurations,
the two levels are
\begin{equation}
\label{eq:agg-levels}
p_c(\vx) \;=\; \mathcal{A}_c\big(\{f_{c,v}(\vx)\}_{v=1}^{V}\big),
\qquad
\hat{p}(\vx) \;=\; \mathcal{R}\big(P_{\mathcal{I}}\big)(\vx).
\end{equation}
We vary $V$, $K$, and the reducer separately, comparing uniform averaging with validation-guided
selection and ensembling \citep{caruana2004ensemble, erickson2020autogluon, erickson2025tabarena}.

We report percentage changes in error relative to the released eight-view default, with
calculation details and confidence intervals in Appendix~\ref{app:aggregation-scaling}.
Elo follows the common protocol in Section~\ref{sec:setup}.

\paragraph{Native-view scaling.}
Increasing native views beyond the default $V=8$ yields diminishing returns (Figure~\ref{fig:agg-canonical}A). Relative to this default, $V=16$ reduces error by $0.48\%$ and $V=32$ by $0.60\%$ ($+21$ Elo). Increasing to $V=64$ provides no clear improvement over $V=32$, which suggests that the gains from native-view averaging have largely saturated. Full results appear in Appendix~\ref{app:aggregation-scaling}.

\paragraph{Aggregation across configurations.}
We next ask whether varying inference configurations provides further gains beyond increasing
native views, and how these gains depend on the reducer. We construct a pool of 96 configurations
from TabArena's tuned-TabPFN search space, adapted to TabPFN-3, and evaluate each with eight
native views at test time.

Using the same member predictions, we compare three reducers. Uniform averaging gives all
members equal weight. Best-single selects the member with the lowest validation loss.
Greedy selection \citep{caruana2004ensemble} builds an ensemble by repeatedly adding the member
that minimizes the updated ensemble's validation loss. Both selection methods use out-of-fold
predictions on the training data. Appendix~\ref{app:aggregation-reducers} and
Algorithm~\ref{alg:aggregation} give the definitions and full procedure.

The same pool produces different outcomes depending on the reducer (Figure~\ref{fig:agg-canonical}B).
A randomly chosen configuration performs worse than the released default on average.
Uniformly averaging all 96 configurations yields a relative error change of $+3.68\%$
(95\% CI: $[+0.02\%, +8.86\%]$). Best-single yields $-1.25\%$
(95\% CI: $[-3.41\%, +0.07\%]$), with an interval that includes zero. Greedy selection
yields $-2.44\%$ (95\% CI: $[-4.75\%, -0.80\%]$; $+67$ Elo), achieving a larger
estimated reduction than best-single. Expanding the pool alone
therefore does not ensure a gain; how its predictions are selected and combined matters.

\begin{figure}[!t]
\centering
\includegraphics[width=0.9\linewidth,trim=0 7bp 0 0,clip]{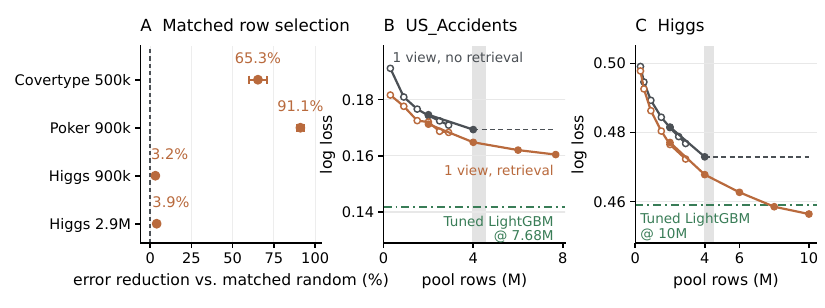}
\caption{Row curation on frozen, single-view TabPFN-3.
\textbf{A}: error reduction of attention-selected versus matched random rows;
95\% paired $t$ intervals over three seeds.
\textbf{B/C}: five-seed mean log loss across source-pool sizes.
Open and filled points use separate splits from 3M-row subsets and full tables, respectively.
Curve labels in B also apply to C. Dashed lines show the no-retrieval loss
at 4M rows; shading marks the 4.0--4.5M memory boundary on a single 80\,GB GPU.}
\label{fig:row-curation}
\end{figure}

We then examine how these comparisons change as the configuration pool grows. For each reducer,
we evaluate random subsets of the same 96 configurations and average over draws at each size
(Figure~\ref{fig:agg-canonical}B). Average error decreases with larger subsets for all three
reducers. Uniform averaging remains worse than the default, while best-single's point estimate
is better than the default from $K=16$ onward. Greedy selection achieves a larger estimated
error reduction than best-single and performs best with the full $K=96$ pool.

\vspace{-2mm}
\paragraph{Dependence on validation data.}
Since greedy selection uses validation loss to construct the ensemble, we next examine how its
gains depend on the amount of validation data available. We hold the 96 configurations and their
predictions fixed and vary the fraction of out-of-fold examples used to guide selection. The error
grows as fewer examples are available: using all examples reduces error by $2.4\%$,
whereas using one eighth yields an estimated error $0.4\%$ above the default. Thus, the gains from
greedy aggregation depend on both the configuration pool and the validation data available for
selection. Full results are provided in Appendix~\ref{app:aggregation-validation}.

%% file: context_construction.tex
\vspace{-3mm}
\section{Context construction: curation and expansion}
\label{sec:context}
We investigate two directions for improving predictions through context construction: curating
existing rows or columns to form the context, and expanding it with synthetic rows or features.
With the backbone frozen, we examine when these changes improve predictions over the original
context.

\vspace{-3mm}
\paragraph{Column curation.}
TabPFN-3 already handles wide tables through feature subsampling and supports importance-guided
selection \citep{priorlabs2026tabpfn3}. We evaluate when such selection improves predictions over
default sampling. We use TabPFN's selector, which fits LightGBM on each training split and ranks columns by total split gain (Appendix~\ref{app:column-curation}).
We fix the top 200 columns for classification or 500 for regression across all eight views,
matching the default per-view counts.

Across seven wide tables, the effects vary substantially. Selection reduces error by $30$--$67\%$
relative to the default on the genomics, mass-spectrometry, and text datasets, but slightly
increases error on three of the four molecular or image datasets. The native importance-based
sampler exhibits a similar pattern (Table~\ref{tab:column-grid}). In a separate comparison that
fixes 500 columns across all eight views, importance-based selection reduces error by $66\%$
relative to random selection on amazon-reviews, showing that the choice of columns matters
even at the same column budget (Table~\ref{tab:column-curation}).

Column selection may be particularly useful when the model must identify relevant information
among many features from only a few context examples. We test this hypothesis by varying the
number of context rows on four wide datasets with relatively large training pools and limited
gains from column selection in the initial comparison. Gains increase at small context sizes
on three datasets, while CIFAR\_10 shows little change (Appendix~\ref{app:column-curation}).

\paragraph{Attention-guided row selection.}
Row selection allows the context to be tailored to the queries being predicted. We investigate
whether this improves predictions over using all available training rows, even when the full
context fits in memory. We adapt the attention-guided retrieval pipeline of LimiX \citep{zhang2025limix}
to frozen TabPFN-3. We rank training rows by query-to-context attention, group queries with
similar retrieved sets, and predict each group using the union of its retrieved rows
(Algorithm~\ref{alg:model-aware-retrieval}).

On 15 TabArena datasets, retrieval has mixed effects under the eight-view default: two datasets
improve and two worsen, with corrected confidence intervals excluding zero (Appendix~\ref{app:row-curation}).
We then evaluate five larger tables using a single native view. At the evaluated sizes where
the full context fits in memory, attention-guided retrieval lowers log loss on four of five
tables (Table~\ref{tab:row-construction}).

To test whether the selected rows matter beyond reducing context size, we replace them with
uniform random samples of the same size, keeping query groups and the refitting procedure fixed.
Attention-selected contexts reduce log loss relative to these matched random contexts by
$65\%$ on Covertype, $91\%$ on Poker, and $3$--$4\%$ on Higgs at two source-pool sizes
(Figure~\ref{fig:row-curation}A). In all four settings, attention-selected contexts outperform
the full-context baseline, whereas matched random contexts perform worse. These comparisons
indicate that the gains depend on which rows are selected, rather than context reduction alone.

\vspace{-1mm}
\paragraph{Beyond the memory limit.}
Retrieval also offers a way to use training pools too large to fit into a single context.
In our setup, full-context inference fits 4M training rows on one 80\,GB H100 GPU but
runs out of memory at 4.5M. We score attention within blocks and combine the retrieved rows,
enabling retrieval from pools of up to 10M rows (Appendix~\ref{app:row-curation}). Compared with
full-context inference at 4M rows, retrieval reduces log loss by $3.5\%$ on Higgs with a 10M-row
pool and $5.2\%$ on US\_Accidents with a 7.68M-row pool, at $3.9$ and $2.1$ times the baseline's
measured cost, respectively (Figure~\ref{fig:row-curation}B/C). Within the full-table splits on Higgs, mean log loss decreases
as the source pool grows to 10M rows.
At the largest evaluated pools, retrieval has lower mean log loss than validation-tuned
LightGBM \citep{ke2017lightgbm} on Higgs, but higher on US\_Accidents
(Appendix~\ref{app:row-lightgbm}).

\vspace{-1mm}
\paragraph{A retrieval failure.}
On COMET\_MC with 900k training rows, attention-guided selection raises log loss from $0.0151$
with the full context to $0.2302$. To investigate this failure, we replace the selected rows with
random samples of the same size, while keeping query groups and refitting fixed. This brings
log loss back to $0.0158$, indicating that context size alone does not explain the degradation.

Inspecting the retrieved contexts reveals that many queries select the same small set of rows.
Retrieval also changes the class composition of the context. The rare class makes up only
$1.7\%$ of the training pool, but its share within the retrieved contexts has a median of
$40.3\%$. We replace some retrieved rows to match the training pool's class proportions,
keeping the number of context rows fixed. This reduces log loss from $0.2302$ to $0.1270$,
still well above the full-context value of $0.0151$, while AUC falls from $0.950$ to $0.569$.
Thus, this adjustment does not resolve the failure (Appendix~\ref{app:comet-diagnostic}).
Together with the gains on other datasets, this failure suggests that the benefits of context
selection depend on how the selection rule interacts with the characteristics of the data.

\vspace{-1mm}
\paragraph{Context expansion.}
Context expansion asks whether adding constructed rows or features can improve predictions from
the frozen backbone. We evaluate perturbations and interpolations of training rows, query rows
augmented with model-predicted labels, and additional features derived from training-label
statistics or out-of-fold predictions. These constructions yield occasional improvements but
no consistent gains across the evaluated datasets.

We also investigate whether optimizing the added rows makes expansion more effective. We optimize
synthetic rows directly or generate them with an MLP trained through the frozen backbone. In the
distillation experiments, the synthetic rows are optimized to support predictions on real
training examples on their own, then appended to the original context at inference. These
learned additions also fail to yield consistent improvements. The tested expansion strategies
therefore provide no reliable gains over the original context
(Appendix~\ref{app:context-generation}).

%% file: analysis.tex
\vspace{-2mm}
\section{Composition, cost, and variation across tasks}
\label{sec:analysis}
We examine whether adaptation and aggregation provide further gains when combined, what
computation these gains require, and whether they can be anticipated for a new task.
Elo follows the protocol in Section~\ref{sec:setup}; Table~\ref{tab:cost} lists the
evaluated recipes.

\begin{figure}[!htbp]
\centering
\includegraphics[width=\linewidth]{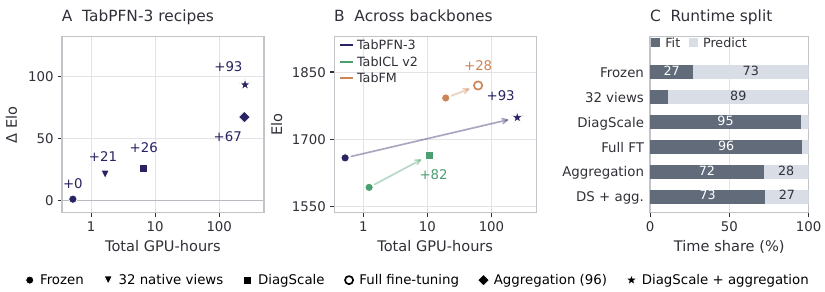}
\caption{Performance and runtime on TabArena.
(A) TabPFN-3 recipes.
(B) Each backbone's frozen model and highest-Elo evaluated recipe, connected by an arrow.
Annotations show $\Delta$ Elo relative to the corresponding reference baseline
within each fit.
(C) Fitting and prediction time shares for TabPFN-3 recipes.
Runtime includes fitting and prediction across 51 datasets
(Appendix~\ref{app:runtime-accounting}).}
\vspace{-3mm}
\label{fig:cost-performance}
\end{figure}

\vspace{-2mm}
\paragraph{Composition.}
Combining DiagScale with 96-configuration greedy aggregation improves on either intervention
alone on TabPFN-3 (Figure~\ref{fig:cost-performance}A). We first adapt TabPFN-3 with DiagScale,
then use the adapted model across all 96 configurations for greedy aggregation. The combination
gains $93$ Elo over the published TabPFN-3 default and reduces error by $2.35\%$ relative to
DiagScale alone and $0.64\%$ relative to aggregation alone (Appendix~\ref{app:composition}).

\vspace{-2mm}
\paragraph{Performance and cost.}
We compare total fitting and prediction time across the same 51 datasets.
On TabPFN-3, 32 native views gain $21$ Elo at 1.66 GPU-hours, while DiagScale gains $26$ Elo
at 6.61 GPU-hours. Aggregation gains $67$ Elo at 247.5 GPU-hours. Combining it with DiagScale
raises the gain to $93$ Elo at 253.1 GPU-hours, a $2.3\%$ cost increase over aggregation alone
(Figure~\ref{fig:cost-performance}A).


The methods differ in where they spend computation (Figure~\ref{fig:cost-performance}C). Increasing native views mainly adds prediction cost, while adaptation spends most of its time fitting. DiagScale and full fine-tuning have similar fitting times despite their very different trainable parameter counts, while prediction remains close to the frozen default. Configuration aggregation spends most of its time generating out-of-fold member predictions for selection.

Frozen TabFM achieves a higher Elo than TabPFN-3 with DiagScale and aggregation, at a lower
total cost (19.3 vs.\ 253.1 GPU-hours; Figure~\ref{fig:cost-performance}B). This comparison
highlights the importance of backbone choice when allocating task-specific compute.
Table~\ref{tab:cost} gives the full runtime measurements.

\paragraph{Variation and predictability of gains.}
Benchmark-level gains are unevenly distributed across datasets. Across seven method--backbone
combinations involving adaptation or aggregation, the five largest positive per-dataset reductions account for
$46$--$70\%$ of the sum of positive reductions. This variation raises the question of whether
gains can be anticipated for a new dataset. For each combination, we examine 21 predictors
(dataset size, feature statistics, disagreement among predictions, etc.). Models combining the
tested predictors achieve at best a cross-validated $R^2$ of $0.012$ when predicting per-dataset
error reductions, indicating limited accuracy in predicting gain magnitudes
(Appendix~\ref{app:predicting-gains}).

%% file: discussion.tex
\vspace{-1mm}
\section{Discussion and conclusion}
\label{sec:discussion}
We systematically study test-time computation for strong tabular foundation models by separating
adaptation, aggregation, and context construction. DiagScale shows that a small set of similarity
updates can capture much of the fine-tuning gain across three backbones. On TabPFN-3, selective
aggregation provides additional gains and combines effectively with DiagScale. Context selection
yields substantial improvements on some datasets, whereas the expansion strategies we tested
provide no consistent benefit.

These results highlight the value of examining individual design choices within a prediction
pipeline. Increasing native views yields diminishing returns, while changing configurations and
selecting among their predictions provides further gains. Uniform averaging and greedy selection
yield different outcomes from the same pool. Similarly, matched row-selection controls show that the choice of context rows can determine whether selection improves or degrades prediction. These comparisons help distinguish the contribution of each operation from the performance of the full pipeline.

Choosing an appropriate test-time compute recipe for a new dataset remains an open problem. Aggregate benchmark gains do not reliably predict whether a particular recipe will benefit an unseen dataset, and we find that the predictors we tested provide limited guidance at the dataset level. Future work should investigate how model--data interactions determine the effectiveness of different test-time compute strategies and use these signals to guide their selection and composition.

%% file: appendix_evaluation.tex
\section{Evaluation and reproducibility details}
\label{app:evaluation}

\subsection{Backbones and model implementations}
\label{app:backbones}

We use three independently pretrained in-context tabular FMs: the open-source base checkpoint of
TabPFN-3 \citep{priorlabs2026tabpfn3} with 53.2--58.3M parameters, which is our primary backbone; TabICL v2
\citep{qu2026tabiclv2} with 27.6--28.5M parameters; and TabFM \citep{google2026tabfm} with 1.64--1.65B parameters,
released under a non-commercial licence for research use. TFM experiments run on a single 80\,GB
H100; the LightGBM row references use CPUs. TabPFN-3 and TabFM are evaluated in bfloat16; TabICL v2 is evaluated with automatic mixed
precision disabled.

Base evaluations use the released packages; adaptation uses project forks of TabICL and TabFM
with parameter-efficient hooks. Package versions, fork commits, and environment specifications
accompany the code.

For Elo, the baselines are the published TabPFN-3 and TabICLv2 reference entries and the frozen
TabFM reproduction of Appendix~\ref{app:tabfm-reproduction}. Paired-error comparisons use the
reproduced frozen configurations with matching runner and inference settings.
These baselines are not identical: reproduced frozen TabICL v2 scores about 9 Elo above its
published default, so Elo gains relative to that default also include the inference-setting
difference. Adapter parameter counts and
fine-tuning hyperparameters are given with the adaptation methods in
Appendices~\ref{app:adaptation-methods}--\ref{app:adaptation-sweeps}.

\subsection{TabFM reproduction}
\label{app:tabfm-reproduction}

TabFM's released TabArena reference results were produced with the JAX backend on TPU with device sharding. We
reproduce them with the released PyTorch backend on one 80\,GB H100. The pretrained weights, the
default preset of 32 augmented views, logit-mean blending, \texttt{n\_estimators}${}=32$, serial
per-view evaluation, the softmax temperature, and the normalization methods are taken from the
released implementation. Our reproduction differs from the reference in the following respects.

The backend is PyTorch rather than JAX and runs on one GPU without sharding. The backbone runs in
bfloat16 rather than the backend's fp32 default, which halves activation memory and allows wide and
large tables to run at full context. TF32 matmul is enabled. Tall-and-wide tables receive a
conditional feature cap: a table keeps the estimator's default of 500 features when
$\text{train rows} \times \min(\text{features}, 500) \le 5$M cells, and otherwise falls back to 100
features. This changes the feature count on APSFailure, kddcup09\_appetency, and
Amazon\_employee\_access; the other 48 datasets retain the reference's setting.
Finally, our runner wraps the estimator in the benchmark's model interface, which applies
the benchmark's standard preprocessing before the estimator sees the data, where the reference ran
the standalone estimator.

TabFM was released after the TabArena snapshot and has no official leaderboard entry, so the
comparison is against the released per-dataset reference numbers rather than an official score.
Across the 51 datasets the median absolute relative difference in task error is 0.22\%, with 47
within 5\% and 42 within 2\%, and our run matches or improves on the reference on 34.
Table~\ref{tab:tabfm-repro} gives the per-dataset comparison.

{\small
\setlength{\tabcolsep}{6pt}
\begin{longtable}{@{}lrrr@{}}
\caption{Single-H100 reproduction of TabFM. Per-dataset error for the TPU/JAX reference against our
bfloat16 H100 run on all 51 TabArena datasets, with
$\Delta = (\text{ours} - \text{reference})/\text{reference}$, so a negative value means our run has
the lower error. All quantities are errors, lower is better.}
\label{tab:tabfm-repro}\\
\toprule
\textbf{Dataset} & \textbf{Reference (TPU/JAX)} & \textbf{Ours (bf16, H100)} & \textbf{$\Delta$\%} \\
\midrule
\endfirsthead
\toprule
\textbf{Dataset} & \textbf{Reference (TPU/JAX)} & \textbf{Ours (bf16, H100)} & \textbf{$\Delta$\%} \\
\midrule
\endhead
\midrule
\multicolumn{4}{r@{}}{\footnotesize\emph{continued on next page}}\\
\endfoot
\bottomrule
\endlastfoot
\multicolumn{4}{@{}l}{\emph{Binary --- reported as $1\!-\!$AUC} (30 datasets)}\\
Amazon\_employee\_access & 0.1416 & 0.1419 & $+0.24$ \\
APSFailure & 0.007387 & 0.005329 & $-27.85$ \\
bank-marketing & 0.2309 & 0.2306 & $-0.13$ \\
Bank\_Customer\_Churn & 0.1229 & 0.1233 & $+0.30$ \\
Bioresponse & 0.1178 & 0.1136 & $-3.55$ \\
blood-transfusion-service-center & 0.2442 & 0.2442 & $-0.00$ \\
churn & 0.06743 & 0.06663 & $-1.19$ \\
coil2000\_insurance\_policies & 0.2124 & 0.2122 & $-0.13$ \\
credit-g & 0.1942 & 0.1942 & $+0.00$ \\
credit\_card\_clients\_default & 0.2068 & 0.2069 & $+0.04$ \\
customer\_satisfaction\_in\_airline & 0.003939 & 0.003793 & $-3.71$ \\
diabetes & 0.158 & 0.1579 & $-0.04$ \\
Diabetes130US & 0.3292 & 0.3299 & $+0.23$ \\
E-CommereShippingData & 0.2576 & 0.2574 & $-0.07$ \\
Fitness\_Club & 0.1789 & 0.1787 & $-0.10$ \\
GiveMeSomeCredit & 0.1313 & 0.1312 & $-0.07$ \\
hazelnut-spread-contaminant-detection & 0.002297 & 0.002257 & $-1.73$ \\
heloc & 0.1977 & 0.1974 & $-0.19$ \\
HR\_Analytics\_Job\_Change\_of\_Data\_Scientists & 0.1938 & 0.1933 & $-0.22$ \\
in\_vehicle\_coupon\_recommendation & 0.1512 & 0.1526 & $+0.92$ \\
Is-this-a-good-customer & 0.2469 & 0.2464 & $-0.18$ \\
jm1 & 0.2106 & 0.2107 & $+0.03$ \\
kddcup09\_appetency & 0.1559 & 0.1596 & $+2.37$ \\
Marketing\_Campaign & 0.07298 & 0.07356 & $+0.80$ \\
NATICUSdroid & 0.01136 & 0.01139 & $+0.29$ \\
online\_shoppers\_intention & 0.06036 & 0.06007 & $-0.48$ \\
polish\_companies\_bankruptcy & 0.005806 & 0.004699 & $-19.06$ \\
qsar-biodeg & 0.05805 & 0.05804 & $-0.02$ \\
seismic-bumps & 0.206 & 0.2061 & $+0.02$ \\
taiwanese\_bankruptcy\_prediction & 0.0503 & 0.04909 & $-2.39$ \\
\midrule
\multicolumn{4}{@{}l}{\emph{Multiclass --- log loss} (8 datasets)}\\
anneal & 0.01609 & 0.0132 & $-17.96$ \\
hiva\_agnostic & 0.206 & 0.1774 & $-13.88$ \\
maternal\_health\_risk & 0.3711 & 0.3706 & $-0.12$ \\
MIC & 0.4289 & 0.4306 & $+0.40$ \\
SDSS17 & 0.06969 & 0.0693 & $-0.56$ \\
splice & 0.09589 & 0.09596 & $+0.07$ \\
students\_dropout\_and\_academic\_success & 0.5072 & 0.5078 & $+0.13$ \\
website\_phishing & 0.2108 & 0.2102 & $-0.29$ \\
\midrule
\multicolumn{4}{@{}l}{\emph{Regression --- RMSE} (13 datasets)}\\
airfoil\_self\_noise & 1.071 & 1.072 & $+0.07$ \\
Another-Dataset-on-used-Fiat-500 & 703.6 & 703.4 & $-0.04$ \\
concrete\_compressive\_strength & 3.986 & 3.97 & $-0.40$ \\
diamonds & 499.7 & 496.5 & $-0.64$ \\
Food\_Delivery\_Time & 7.33 & 7.333 & $+0.04$ \\
healthcare\_insurance\_expenses & 4420.1 & 4407.6 & $-0.28$ \\
houses & 0.1816 & 0.1815 & $-0.08$ \\
miami\_housing & 74223.7 & 74051.8 & $-0.23$ \\
physiochemical\_protein & 2.888 & 2.886 & $-0.07$ \\
QSAR-TID-11 & 0.7548 & 0.7357 & $-2.52$ \\
QSAR\_fish\_toxicity & 0.8535 & 0.8537 & $+0.02$ \\
superconductivity & 8.952 & 8.941 & $-0.12$ \\
wine\_quality & 0.5864 & 0.5864 & $-0.00$ \\
\end{longtable}
}

\subsection{Datasets, splits, and evaluation coverage}
\label{app:datasets}
\label{app:evidence-tiers}

\paragraph{TabArena.}
The benchmark comprises 51 curated datasets: 30 binary classification, 8 multiclass, and 13
regression. They range from 748 to 150{,}000 rows (median 6{,}497) and from 4 to 1{,}776 features
(median 17). We use three-fold cross-validation, with 10 repeats for the 17 datasets below
2{,}500 rows and 3 repeats for the remaining 34. Each dataset--repeat--fold evaluation is a
\emph{cell}: the full protocol has $17\times30+34\times9=816$ cells, and the first repeat has
$51\times3=153$. All reported Elo ratings use the full protocol. Controlled subsets are identified
in their respective tables or captions. Per-dataset identifiers, sizes, task types, and metrics
follow the benchmark.

\paragraph{External tables.}
Row curation is additionally studied on five raw OpenML tables larger than the TabArena datasets
(Table~\ref{tab:external-datasets}). For each of five seeds we draw a 50{,}000-row test set and
sample context pools of varying sizes from the remaining rows. Methods within a comparison share
the same seed-specific partition. These are study-specific random partitions, not published
OpenML task splits. Table~\ref{tab:external-datasets} distinguishes total source rows from the
largest evaluated pool; final context sizes are reported with each method in
Appendix~\ref{app:row-curation}. Wide-table datasets and their splits are specified in
Appendix~\ref{app:column-curation}. External-table results use task errors rather than Elo.

\begin{table}[t]
\centering
\small
\caption{External OpenML tables used for row curation. Predictors and task type are the effective
inputs seen by the model. \emph{Largest pool} is the biggest context pool evaluated. The US\_Accidents and COMET pools use all remaining rows after the 50,000-row holdout;
Higgs is capped at 10M.}
\label{tab:external-datasets}
\begin{tabular}{@{}lrrrlr@{}}
\toprule
\textbf{Dataset} & \textbf{OpenML ID} & \textbf{Total rows} & \textbf{Predictors} & \textbf{Task} & \textbf{Largest pool} \\
\midrule
Higgs         & 45570 & 11{,}000{,}000 & 28 & binary          & 10{,}000{,}000 \\
US\_Accidents & 46650 &  7{,}728{,}394 & 43 & 4-class         &  7{,}678{,}394 \\
COMET\_MC     &  5889 &  7{,}619{,}400 &  4 & binary          &  7{,}569{,}400 \\
Poker-Hand    &  1567 &  1{,}025{,}009 & 10 & 10-class        &     900{,}000 \\
Covertype     &  1596 &    581{,}012   & 54 & 7-class         &     500{,}000 \\
\bottomrule
\end{tabular}
\end{table}

\subsection{Metrics and reference models}
\label{app:evaluation-grid}

\paragraph{Task metrics.}
Tasks are scored with TabArena's official metrics \citep{erickson2025tabarena}: ROC AUC for binary
classification, log loss for multiclass classification, and RMSE for regression.
We use TabArena's metric implementation, including its probability handling for log loss.
All comparisons use a lower-is-better error $e$: $1-\operatorname{AUC}$ for binary tasks,
log loss for multiclass tasks, and RMSE for regression. The external context studies state their
metrics separately in Appendix~\ref{app:context}; the large-row studies use log loss.

\paragraph{Elo and the comparison field.}
Let $e_{m,d,c}$ be method $m$'s error in cell $c$ of dataset $d$, and let $C_d$ be the number
of cells for that dataset. Each pair of methods is compared within each cell: the lower error
wins, and an exact tie contributes half a win to each. TabArena fits all ratings jointly under
the Bradley--Terry model
\begin{equation}
P(m\text{ beats }j)=\frac{1}{1+10^{(R_j-R_m)/400}}.
\end{equation}
The fit maximizes the weighted pairwise log likelihood, with cell weight $1/C_d$ so that each
dataset contributes equally to each method pair. We use TabArena's logistic-regression solver
and shift all fitted ratings to set RandomForest (default) to 1000. Higher Elo is better;
the displayed ratings are rounded to integers.
All reported Elo results, including the adaptation menu, cross-backbone comparisons,
aggregation, and composition, use the same 68 reference entries
(Table~\ref{tab:reference-models}): the 67 TabArena entries plus frozen TabFM.
The reference field includes tuned classical learners, gradient-boosted trees, tabular foundation
models, and AutoGluon. Each candidate is inserted separately, yielding a 69-entry fit.
Within that fit, $\Delta$ Elo is measured against the published TabPFN-3 or TabICLv2 default,
or against frozen TabFM for TabFM adaptations. We compute the difference before rounding it
to an integer, and use unrounded differences to select learning rates in the adaptation-menu sweep.
The reference predictions are fixed, but their fitted ratings may change with the candidate;
we therefore use the baseline rating from that candidate's fit for every difference.
Paired-error intervals are defined separately below.

\paragraph{Normalized score.}
The \emph{Score} column in the adaptation sweep is TabArena's normalized score. First average
errors over cells within each dataset, $\bar e_{m,d}=C_d^{-1}\sum_c e_{m,d,c}$.
Within the comparison field, let $a_d=\min_m\bar e_{m,d}$ and
$b_d=\operatorname{median}_m\bar e_{m,d}$. The reported score is
\begin{equation}
\operatorname{Score}_m=\frac{1}{D}\sum_{d=1}^{D}
\left[1-\operatorname{clip}_{[0,1]}
\left(\frac{\bar e_{m,d}-a_d}{\max(b_d-a_d,10^{-5})}\right)\right],
\end{equation}
where $D$ is the number of datasets. Higher is better, and the score lies in $[0,1]$.
The best method on a dataset receives 1; errors at least $\max(b_d-a_d,10^{-5})$ above the
best receive 0. Like Elo, this score depends on the comparison field.

\paragraph{Paired error changes and uncertainty.}
For a fixed reference method $b$, the benchmark-level paired summary first averages log error
ratios within each dataset and then weights datasets equally:
\begin{equation}
\ell_d=\frac{1}{C_d}\sum_c\log\frac{e_{m,d,c}}{e_{b,d,c}},
\qquad
\Delta_{\rm error}(\%)=100\left[\exp\left(\frac{1}{D}\sum_d\ell_d\right)-1\right].
\end{equation}
Negative values indicate improvement. We form 95\% percentile intervals from 10{,}000 bootstrap
resamples of datasets, retaining all cells within each sampled dataset and applying the same
exponential transformation to the interval endpoints. These intervals condition on the selected
recipes. In these summaries, \emph{Wins} counts datasets with $\ell_d<0$.
The adaptation tables also summarize per-dataset relative changes
$r_d=100(\bar e_{m,d}/\bar e_{b,d}-1)$, where $\bar e$ averages cell errors within a dataset;
their mean is the arithmetic average of $r_d$, with negative values indicating improvement.
The per-dataset gain-distribution analyses instead use
$100(1-\bar e_{m,d}/\bar e_{b,d})$, for which positive values indicate improvement
(Appendix~\ref{app:gain-concentration}). This ratio of arithmetic mean errors and the mean log
ratio are different summaries, so their win counts can differ.

\paragraph{Selection and information sources.}
Task-specific choices, including checkpoints and reducer weights, use validation or out-of-fold
predictions within the training split. Study-level learning rates are selected using aggregate
test performance on the same benchmark used to report results, rather than on held-out tasks;
both TabFM methods use a five-dataset subset, and each selected rate is shared across datasets
(Appendix~\ref{app:adaptation-sweeps}).

\paragraph{Reference entries.}
Table~\ref{tab:reference-models} lists the 68 fixed references: 67 TabArena entries and
our frozen TabFM reproduction. The count refers to benchmark entries, including separate
default, tuned, and tuned + ensembled variants, rather than model families. We retain
TabArena's published variant names and reference results \citep{erickson2025tabarena};
reused TabArena and TabRepo predictions \citep{salinas2023tabrepo} supply the reference field.
Missing reference results are filled using the benchmark's default RandomForest fallback;
the table gives the dataset-weighted imputation percentage. All evaluated recipes and the
frozen TabFM reference have complete results and zero imputation.

{\small
\begin{longtable}{@{}lr@{}}
\caption{The 68 reference entries used in every Elo comparison. Imputed (\%) records benchmark
fallback results. The published TabPFN-3 and TabICLv2 defaults are distinct from our reproduced frozen models.}
\label{tab:reference-models}\\
\toprule
\textbf{Reference entry} & \textbf{Imputed (\%)} \\
\midrule
\endfirsthead
\multicolumn{2}{l}{\tablename~\thetable{} (continued)}\\
\toprule
\textbf{Reference entry} & \textbf{Imputed (\%)} \\
\midrule
\endhead
\midrule
\multicolumn{2}{r@{}}{\footnotesize\emph{continued on next page}}\\
\endfoot
\bottomrule
\endlastfoot
TabFM (frozen reproduction) & 0 \\
AutoGluon 1.5 (extreme, 4h) & 0 \\
TabPFN-3 (default) & 0 \\
TabPFN-2.6 (default) & 0 \\
TabICLv2 (default) & 0 \\
RealTabPFN-2.5 (tuned + ensembled) & 0 \\
RealTabPFN-2.5 (tuned) & 0 \\
RealTabPFN-2.5 (default) & 0 \\
AutoGluon 1.4 (best, 4h) & 0 \\
RealMLP (tuned + ensembled) & 0 \\
TabDPT (tuned + ensembled) & 0 \\
TabM (tuned + ensembled) & 0 \\
LightGBM (tuned + ensembled) & 0 \\
RealMLP (tuned) & 0 \\
CatBoost (tuned + ensembled) & 0 \\
CatBoost (tuned) & 0 \\
TabDPT (tuned) & 0 \\
iLTM (tuned + ensembled) & 0 \\
TabM (tuned) & 0 \\
ModernNCA (tuned + ensembled) & 0 \\
LightGBM (tuned) & 0 \\
XGBoost (tuned + ensembled) & 0 \\
CatBoost (default) & 0 \\
ModernNCA (tuned) & 0 \\
XGBoost (tuned) & 0 \\
xRFM (tuned + ensembled) & 0 \\
TabPFNv2 (tuned + ensembled) & 35.29 \\
Mitra (default) & 35.29 \\
TabDPT (default) & 0 \\
TabICL (default) & 29.41 \\
xRFM (tuned) & 0 \\
TabM (default) & 0 \\
iLTM (tuned) & 0 \\
TabPFNv2 (tuned) & 35.29 \\
TorchMLP (tuned + ensembled) & 0 \\
EBM (tuned + ensembled) & 0 \\
TabPFNv2 (default) & 35.29 \\
ModernNCA (default) & 0 \\
EBM (tuned) & 0 \\
RealMLP (default) & 0 \\
XGBoost (default) & 0 \\
TorchMLP (tuned) & 0 \\
FastaiMLP (tuned + ensembled) & 0 \\
EBM (default) & 0 \\
ExtraTrees (tuned + ensembled) & 0 \\
LightGBM (default) & 0 \\
RandomForest (tuned + ensembled) & 0 \\
ExtraTrees (tuned) & 0 \\
FastaiMLP (tuned) & 0 \\
RandomForest (tuned) & 0 \\
TabSTAR (tuned) & 0 \\
TabSTAR (tuned + ensembled) & 0 \\
iLTM (default) & 0 \\
PerpetualBooster (tuned + ensembled) & 0 \\
TorchMLP (default) & 0 \\
PerpetualBooster (tuned) & 0 \\
xRFM (default) & 0 \\
FastaiMLP (default) & 0 \\
TabSTAR (default) & 0 \\
RandomForest (default) & 0 \\
KNN (tuned + ensembled) & 0 \\
ExtraTrees (default) & 0 \\
Linear (tuned + ensembled) & 0 \\
Linear (tuned) & 0 \\
PerpetualBooster (default) & 0 \\
KNN (tuned) & 0 \\
Linear (default) & 0 \\
KNN (default) & 0 \\
\end{longtable}
}

%% file: appendix_adaptation.tex
\section{Adaptation: implementation and additional results}
\label{app:adaptation}

\subsection{Adaptation methods and parameterizations}
\label{app:adaptation-methods}

The sweep contains three families. \emph{Full fine-tuning} updates all trainable weights. \emph{Selective}
settings freeze the backbone except for a named subset of modules: sets
of projections within attention (query--key, value--out, the attention out-projection alone), whole
sublayer types (all attention, all MLP), depth ranges (the first one to three ICL blocks with the
embeddings, the middle three blocks, four blocks spread through the stack, the last one to three
blocks with the head), and individual small components (all LayerNorms, all biases, the inducing
vectors, the input/feature/label embeddings, the decoder head, the label encoder, the
softmax-scaling modules). \emph{LoRA} \citep{hu2022lora} inserts low-rank adapters of rank 4, 8, or 16 on every linear
layer. Table~\ref{tab:full-ladder} lists all 25.

\paragraph{DiagScale.}
On TabPFN-3, each softmax-scaling module is replaced by a wrapper that keeps the pretrained
scale network frozen and multiplies its output by $1 + \vdelta$, with $\vdelta$ zero-initialized and
one entry per (head, dimension). All scaling modules are adapted.
On TabICL v2 and TabFM, the multiplier is registered in each attention module and applied to the
final query before the attention dot product, after the backbone's own query transformations.
At $\vdelta=\vzero$, DiagScale reproduces the frozen backbone exactly.
These adapters are kept in float32 and cast to the attention compute dtype in the forward pass.
The evaluated implementations retain the multiplicative adapters during inference.

The trainable parameter count on TabPFN-3 is task-dependent. The backbone builds three scaling
modules in the feature embedder ($3 \times 128 = 384$ entries) and 24 in the ICL blocks
($24 \times 512 = 12{,}288$), and adds one in the many-class decoder ($384$) only when the task type
is classification. DiagScale therefore trains $12{,}672$ parameters on a regression task and
$13{,}056$ on a classification task. Table~\ref{tab:cross-backbone} reports regression
counts for all three backbones; the benchmark uses both versions across its 38 classification
and 13 regression datasets. DiagScale trains $7{,}296$ parameters on TabICL v2 and $53{,}760$
on TabFM for either task type. Table~\ref{tab:checkpoint-parameters} divides
these counts by each task type's complete checkpoint, counting model parameters rather than
only attention weights or an unfrozen subset. Checkpoint buffers are excluded.

\begin{table}[h]
\centering
\small
\caption{Exact checkpoint and DiagScale parameter counts. Classification and regression use
separate checkpoints. Full-model counts include fixed parameters. Percentages are rounded.}
\label{tab:checkpoint-parameters}
\begin{tabular}{@{}llrrr@{}}
\toprule
\textbf{Backbone} & \textbf{Task} & \textbf{Full checkpoint} & \textbf{DiagScale} & \textbf{Fraction (\%)} \\
\midrule
TabPFN-3 & Classification & 53{,}153{,}144 & 13{,}056 & 0.02456 \\
         & Regression & 58{,}274{,}944 & 12{,}672 & 0.02175 \\
TabICL v2 & Classification & 27{,}552{,}258 & 7{,}296 & 0.02648 \\
          & Regression & 28{,}544{,}991 & 7{,}296 & 0.02556 \\
TabFM & Classification & 1{,}639{,}444{,}298 & 53{,}760 & 0.00328 \\
      & Regression & 1{,}647{,}782{,}989 & 53{,}760 & 0.00326 \\
\bottomrule
\end{tabular}
\end{table}

\subsection{Training and model-selection protocol}
\label{app:adaptation-protocol}

Adaptation splits the task's training data into fitting and validation sets.
Training runs for at most 30 epochs with early-stopping patience 8.
Validation selects the epoch; if no epoch improves on the pretrained weights, the run returns those weights
unchanged. Reported performance therefore reflects both parameter updates and validation-based
selection.

Fine-tuning and checkpoint validation use 2 views on all backbones. Deployment uses 8 views on TabPFN-3 and TabICL v2 and 32
on TabFM, matching each backbone's own default. TabPFN-3 and TabFM run in bfloat16; TabICL v2 runs
with automatic mixed precision and FlashAttention-3 both disabled. Weight decay is fixed at $0.01$.

\subsection{Learning-rate selection and sensitivity}
\label{app:adaptation-sweeps}

\paragraph{Shared recipes and task-level validation.}
We use one learning rate per backbone and adaptation method across datasets.
Shared hyperparameter defaults aim to provide reusable configurations while
reducing per-task tuning costs \citep{probst2019tunability,holzmuller2024better}. We report the cross-backbone learning-rate grids and
assess sensitivity to learning-rate choice and dataset partitioning below.

\paragraph{Tested grids.}
Our preliminary experiments suggested that updating more parameters generally required smaller learning rates, motivating a lower learning-rate range for full fine-tuning than for DiagScale. For the cross-backbone comparison, full FT on TabPFN-3 and TabICL v2 uses
$\{10^{-6},3\times10^{-6},10^{-5},3\times10^{-5},10^{-4}\}$; DiagScale uses
$\{10^{-3},3\times10^{-3},10^{-2},3\times10^{-2},10^{-1}\}$.
Each rate is evaluated over the full 816-cell protocol. The selected rates are
$10^{-6}$ for TabPFN-3 full FT, $10^{-5}$ for TabICL v2 full FT, and
$3\times10^{-2}$ for DiagScale on both backbones.

\begin{figure}[!htbp]
\centering
\includegraphics[width=\linewidth]{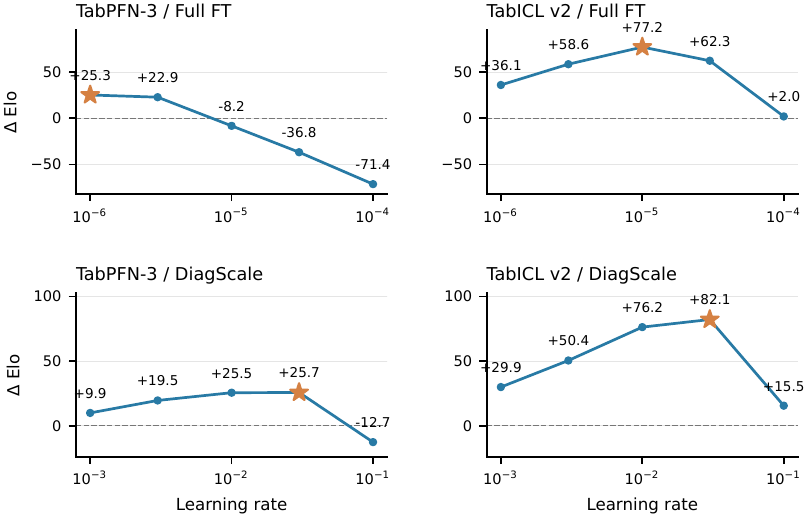}
\caption{Learning-rate sensitivity for the cross-backbone adaptation comparison.
Each point adds one candidate to the same 68 references; $\Delta$ Elo is relative
to the corresponding published default. Stars mark the rates with the highest
unrounded $\Delta$ Elo.
All panels use five learning rates evaluated on the same 51 datasets.}
\label{fig:lr-sensitivity}
\end{figure}

\paragraph{Sensitivity within the tested grids.}
The full-FT and DiagScale curves peak at different learning rates.
For TabPFN-3 and TabICL v2, we select the rate with the highest unrounded
$\Delta$ Elo. TabPFN-3 DiagScale falls below the default at its largest
tested rate. On TabICL v2, both methods improve over the default at all five tested rates.
Across the method-specific learning-rate ranges tested here, DiagScale's
$\Delta$ Elo varies less than full FT's on both backbones, suggesting greater
robustness to learning-rate choice. Restricting updates to dimension-specific
scaling may provide strong regularization, making this reduced sensitivity a
potential practical advantage over full fine-tuning.

We also repeat five-fold cross-validation over datasets with 20 random partitions.
Each fold selects the rate with the highest $\Delta$ Elo on 40 or 41 datasets and evaluates it
on the remaining 10 or 11, keeping all splits of a dataset together. Each repetition combines
the held-out results into one Elo fit (Table~\ref{tab:lr-dataset-cv}).

\begin{table}[!htb]
\centering\small
\caption{Learning-rate selection across datasets. Entries are $\Delta$ Elo against the published
default, with 68 references and one candidate per fit. Global selects one rate using all
51 datasets; five-fold results report the mean and standard deviation across 20 dataset partitions.}
\label{tab:lr-dataset-cv}
\setlength{\tabcolsep}{10pt}
\begin{tabular}{@{}llrr@{}}
\toprule
Backbone & Method & Global & Five-fold (mean $\pm$ SD) \\
\midrule
TabPFN-3 & Full FT & $+25.27$ & $+21.17 \pm 2.17$ \\
TabPFN-3 & DiagScale & $+25.66$ & $+19.90 \pm 2.06$ \\
TabICL v2 & Full FT & $+77.20$ & $+77.20 \pm 0.00$ \\
TabICL v2 & DiagScale & $+82.12$ & $+78.64 \pm 3.64$ \\
\bottomrule
\end{tabular}
\end{table}

Within the tested grids, both methods retain positive $\Delta$ Elo in every partition,
with similar mean Elo on each backbone.





\paragraph{Complete adaptation menu.}
Each of the 25 settings in Table~\ref{tab:full-ladder} is evaluated at
$10^{-6}$, $10^{-5}$, and $10^{-4}$ over the full protocol; the table reports the
rate with the highest aggregate $\Delta$ Elo for each setting. Of these settings, 15 select a grid endpoint:
nine at $10^{-6}$ and six at $10^{-4}$, including SoftScale at the upper endpoint.

{\small
\setlength{\tabcolsep}{2pt}
\begin{longtable}{@{}l>{\raggedright\arraybackslash}p{2.45cm}rrrlrr@{}}
\caption{All 25 adaptation settings on TabPFN-3, each at the global learning rate with the highest
$\Delta$ Elo from $\{10^{-6}, 10^{-5}, 10^{-4}\}$. Each candidate is added separately to the same
68 references; $\Delta$ Elo is relative to the published TabPFN-3 default, and \emph{Score} is
TabArena's normalized score. Error is the mean per-dataset relative change against the frozen
backbone, with a 95\% dataset-bootstrap interval; negative is better. \emph{Wins} counts datasets
with lower error. \emph{Fallback} is the fraction of evaluations where validation selects pretrained
weights. Parameter counts use the regression checkpoint. Rows are ordered by unrounded $\Delta$ Elo.}
\label{tab:full-ladder}\\
\toprule
\textbf{Setting} & \textbf{Updated} & \textbf{Params} & \textbf{$\Delta$ Elo} & \textbf{Score} & \textbf{Error [95\% CI]} & \textbf{Wins} & \textbf{Fallback} \\
\midrule
\endfirsthead
\toprule
\textbf{Setting} & \textbf{Updated} & \textbf{Params} & \textbf{$\Delta$ Elo} & \textbf{Score} & \textbf{Error [95\% CI]} & \textbf{Wins} & \textbf{Fallback} \\
\midrule
\endhead
\bottomrule
\endlastfoot
softscale & query scaling & 1.0M & $+26$ & $0.703$ & $-0.62\%$ $[-1.04\%, -0.24\%]$ & 34/51 & $32\%$ \\
full & all trainable weights & 58.3M & $+25$ & $0.704$ & $-0.64\%$ $[-1.07\%, -0.26\%]$ & 37/51 & $36\%$ \\
qk & query + key projections & 12.9M & $+22$ & $0.696$ & $-0.54\%$ $[-0.95\%, -0.16\%]$ & 33/51 & $35\%$ \\
middle3 & middle ICL blocks 10-12 & 6.4M & $+20$ & $0.696$ & $-0.60\%$ $[-1.08\%, -0.19\%]$ & 36/51 & $34\%$ \\
attn & all attention sublayers & 26.2M & $+20$ & $0.693$ & $-0.33\%$ $[-0.61\%, -0.09\%]$ & 35/51 & $42\%$ \\
mlp & all MLP sublayers & 25.2M & $+19$ & $0.693$ & $-0.37\%$ $[-0.66\%, -0.11\%]$ & 38/51 & $42\%$ \\
vo & value + out projections & 12.9M & $+19$ & $0.693$ & $-0.38\%$ $[-0.68\%, -0.12\%]$ & 39/51 & $39\%$ \\
inducing & inducing vectors & 49{,}152 & $+18$ & $0.690$ & $-0.28\%$ $[-0.57\%, -0.03\%]$ & 35/51 & $36\%$ \\
attnout & attention out-projection & 6.4M & $+17$ & $0.688$ & $-0.60\%$ $[-1.10\%, -0.15\%]$ & 32/51 & $36\%$ \\
emb-first3 & embeddings + ICL blocks 0-2 & 7.7M & $+16$ & $0.689$ & $-0.27\%$ $[-0.50\%, -0.07\%]$ & 34/51 & $41\%$ \\
ends & embeddings + decoder head & 6.9M & $+16$ & $0.688$ & $-0.23\%$ $[-0.42\%, -0.09\%]$ & 28/51 & $44\%$ \\
emb-first1 & embeddings + ICL block 0 & 3.4M & $+16$ & $0.688$ & $-0.24\%$ $[-0.41\%, -0.10\%]$ & 32/51 & $43\%$ \\
spread4 & blocks 0/8/16/23 & 8.6M & $+16$ & $0.691$ & $-0.47\%$ $[-0.87\%, -0.13\%]$ & 33/51 & $37\%$ \\
lora-r16 & LoRA rank 16 & 3.5M & $+15$ & $0.684$ & $-0.48\%$ $[-0.88\%, -0.13\%]$ & 34/51 & $38\%$ \\
lora-r4 & LoRA rank 4 & 877{,}100 & $+15$ & $0.682$ & $-0.70\%$ $[-1.33\%, -0.16\%]$ & 30/51 & $35\%$ \\
emb-first2 & embeddings + ICL blocks 0-1 & 5.5M & $+14$ & $0.688$ & $-0.23\%$ $[-0.42\%, -0.06\%]$ & 34/51 & $44\%$ \\
emb & embeddings & 1.3M & $+14$ & $0.687$ & $-0.23\%$ $[-0.42\%, -0.08\%]$ & 32/51 & $43\%$ \\
lora-r8 & LoRA rank 8 & 1.8M & $+12$ & $0.678$ & $-0.28\%$ $[-0.62\%, +0.01\%]$ & 32/51 & $41\%$ \\
last3-head & last 3 ICL blocks + head & 12.1M & $+11$ & $0.685$ & $-0.26\%$ $[-0.48\%, -0.07\%]$ & 25/51 & $46\%$ \\
ln & all LayerNorms & 28{,}288 & $+10$ & $0.682$ & $-0.08\%$ $[-0.15\%, -0.02\%]$ & 31/51 & $49\%$ \\
last2-head & last 2 ICL blocks + head & 9.9M & $+6$ & $0.678$ & $-0.13\%$ $[-0.28\%, -0.01\%]$ & 22/51 & $49\%$ \\
bitfit & biases & 24{,}504 & $+4$ & $0.677$ & $-0.06\%$ $[-0.14\%, +0.01\%]$ & 27/51 & $53\%$ \\
last1-head & last ICL block + head & 7.8M & $+3$ & $0.676$ & $-0.06\%$ $[-0.14\%, +0.00\%]$ & 19/51 & $54\%$ \\
yenc & y-encoder only & 1{,}280 & $+2$ & $0.677$ & $-0.01\%$ $[-0.07\%, +0.05\%]$ & 23/51 & $56\%$ \\
head & decoder + output head & 5.7M & $+2$ & $0.674$ & $-0.07\%$ $[-0.15\%, +0.00\%]$ & 25/51 & $52\%$ \\
\end{longtable}
}

\paragraph{Learning rates for TabFM.}
Full fine-tuning and DiagScale each use one learning rate selected on the same five-dataset
subset, covering 87 evaluation cells, and applied across all 51 datasets.
The subset comprises in\_vehicle\_coupon\_recommendation, airfoil\_self\_noise,
hazelnut-spread-contaminant-detection, houses, and polish\_companies\_bankruptcy,
chosen for their combined full-FT test gains on TabPFN-3 and TabICL v2.
Both methods select the rate with the highest mean per-cell relative error reduction on this subset.
TabFM uses $10^{-4}$ for full FT and $10^{-2}$ for DiagScale.


\subsection{Complete cross-backbone results}
\label{app:adaptation-results}

\paragraph{Direct comparison with full fine-tuning.}
We pair DiagScale and full fine-tuning on the same 816 cells for each backbone. Let
$\bar\ell_d$ be the within-dataset mean of $\log(e_{\rm DiagScale}/e_{\rm FT})$.
Table~\ref{tab:diagscale-paired} transforms the mean over 51 equally weighted datasets
with $100(\exp(\cdot)-1)$. Its intervals resample datasets 10{,}000 times and are conditional on
the selected learning rates and deployed update-and-selection recipes.

\begin{table}[h]
\centering\small
\caption{Paired error change of DiagScale relative to full fine-tuning; negative is better.
Wins count datasets with $\bar\ell_d<0$. The intervals describe the comparison at the selected
recipes and do not establish statistical equivalence.}
\label{tab:diagscale-paired}
\begin{tabular}{@{}lrrr@{}}
\toprule
\textbf{Backbone} & \textbf{Error change} & \textbf{95\% interval} & \textbf{Wins} \\
\midrule
TabPFN-3 & $-0.091\%$ & $[-0.283\%,+0.084\%]$ & 21/51 \\
TabICL v2 & $-0.458\%$ & $[-1.064\%,-0.016\%]$ & 26/51 \\
TabFM & $+0.102\%$ & $[-0.054\%,+0.286\%]$ & 28/51 \\
\bottomrule
\end{tabular}
\end{table}

\paragraph{Per-dataset distributions.}
Table~\ref{tab:adapt-dist} summarizes per-dataset error changes against the frozen backbone.
Median changes are small, while larger improvements on some datasets contribute to the aggregate
gains (Table~\ref{tab:concentration}). On TabFM, DiagScale lowers mean relative error by $0.29\%$
but slightly increases the median error ($+0.01\%$), improving 25 of 51 datasets.
Dataset size has only weak correlations with improvement ($\rho=-0.15$ to $+0.26$).

\begin{table}[t]
\centering
\small
\caption{Per-dataset relative error change against the frozen backbone, 51 datasets per row, full
protocol; negative is better. \emph{Best} is the largest single
reduction. \emph{Wins} counts datasets with any reduction. These are the deployed arms at fixed learning rates, including checkpoint selection
and fallback.
$\rho$ is the Spearman correlation
between training-set size and the reduction.}
\label{tab:adapt-dist}
\begin{tabular}{@{}llrrrrr@{}}
\toprule
\textbf{Backbone} & \textbf{Method} & \textbf{Median} & \textbf{Mean} & \textbf{Best} & \textbf{Wins} & \textbf{$\rho$(size)} \\
\midrule
TabPFN-3  & full fine-tuning & $-0.10\%$ & $-0.64\%$ & $-5.9\%$ & 37/51 & $+0.13$ \\
TabPFN-3  & DiagScale & $-0.08\%$ & $-0.73\%$ & $-7.8\%$ & 31/51 & $+0.05$ \\
TabICL v2 & full fine-tuning & $-0.34\%$ & $-1.30\%$ & $-10.9\%$ & 35/51 & $+0.26$ \\
TabICL v2 & DiagScale        & $-0.27\%$ & $-1.60\%$ & $-11.4\%$ & 34/51 & $+0.26$ \\
TabFM     & full fine-tuning & $-0.08\%$ & $-0.32\%$ & $-5.8\%$  & 31/51 & $+0.16$ \\
TabFM     & DiagScale        & $+0.01\%$ & $-0.29\%$ & $-5.8\%$  & 25/51 & $-0.15$ \\
\bottomrule
\end{tabular}
\end{table}

\paragraph{Fallback in the evaluated arms.}
Validation selects the pretrained weights in 297/252 of the 816 full-FT/DiagScale evaluations
on TabPFN-3, 225/211 on TabICL v2, and 196/173 on TabFM. These fallback cases are included
in the reported results.

\subsection{Mechanism diagnostics and ablations}
\label{app:adaptation-mechanisms}

The mechanism experiments use three diagnostic datasets: \texttt{houses} (regression),
\texttt{physiochemical\_protein} (regression), and \texttt{polish\_companies\_bankruptcy}
(binary classification). These diagnostics are separate from the full-benchmark comparisons.

\begin{figure}[h]
\centering
\includegraphics[width=.31\linewidth]{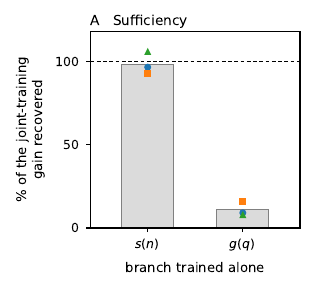}\hfill
\includegraphics[width=.31\linewidth]{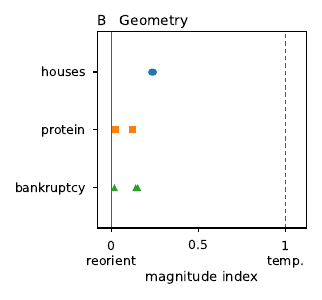}\hfill
\includegraphics[width=.31\linewidth]{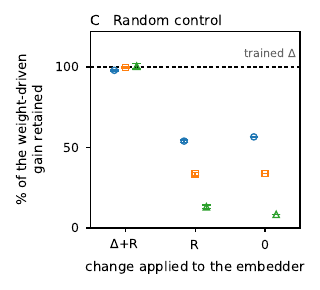}
\caption{SoftScale component diagnostics. Blue circles, orange squares, and green triangles
denote houses, protein, and bankruptcy. A: fraction of the joint-training gain recovered by
separately training only the context-size branch $\vs$ or the query-dependent gate $\vg$;
gray bars show the mean across datasets.
B: each point summarizes one module group: the feature embedder or ICL stack,
plus the classification decoder for bankruptcy. C: feature-embedder controls relative to its
pretrained weights: trained change plus random perturbation ($\Delta+R$), random perturbation
alone ($R$), or reversion ($0$); other scale modules remain trained. Error bars in C show standard
deviations over five draws for $\Delta+R$ and $R$. Gains in A and C are normalized by the
corresponding loss reduction from jointly training both scale branches relative to pretrained
weights along the same inference path.}
\label{fig:diagscale-mech}
\end{figure}

\paragraph{Diagnostic protocol.}
The SoftScale diagnostic runs use learning rate $10^{-4}$ and 30 training epochs. Houses uses fold 0,
repeat 0 (13,760 training rows, 6,000 evaluation rows); protein uses fold 0, repeat 1 (30,486
training rows, 6,000 evaluation rows); bankruptcy uses fold 0, repeat 0 (3,940 training rows,
1,970 evaluation rows). The first two are scored by RMSE and the third by log loss.
Random controls vary five perturbation draws conditional on the checkpoint.
A SoftScale module produces $\tilde{\vq} = \vq \odot \vs(n) \odot \vg(\vq)$, where $n$ is
the number of key tokens in that attention call. The scale network receives $\log n$, and $\vg$
is a query-dependent gate. Thus $n$ is the local attention-context length, rather than one
table-wide row count shared by every module.

\paragraph{Training individual branches.}
We run three independent fine-tunes from the pretrained model: both SoftScale branches,
the context-size branch $\vs$ alone, and the query-dependent gate $\vg$ alone. Each run freezes
all parameters outside its designated branches and selects its checkpoint using inner validation.
Let $L_0$ be the loss with pretrained scale weights along the
fine-tuning inference path, $L_{s,g}$ the loss after training both branches, and $L_b$ the loss
after training only branch $b$. We report $100(L_0-L_b)/(L_0-L_{s,g})$ as the percentage of
the joint-training gain recovered. Training $\vs$ alone recovers $92.8$--$106.2\%$, whereas
training $\vg$ alone recovers $8.1$--$16.0\%$ (Table~\ref{tab:softscale-branch-training}).
A value above $100\%$ means that the restricted run outperforms joint training on that split.

\begin{table}[htbp]
\centering\small
\caption{Separate branch training on the three diagnostic datasets. Entries report the percentage
of the joint SoftScale training gain recovered; training both branches defines $100\%$.
All other parameters remain frozen in each run.}
\label{tab:softscale-branch-training}
\begin{tabular}{@{}lrr@{}}
\toprule
Dataset & Train $\vs$ only & Train $\vg$ only \\
\midrule
Houses & $96.7\%$ & $9.2\%$ \\
Protein & $92.8\%$ & $16.0\%$ \\
Bankruptcy & $106.2\%$ & $8.1\%$ \\
\bottomrule
\end{tabular}
\end{table}

\paragraph{Magnitude against direction.}
For each module we compare the learned change $\Delta\vb$ in the scale vector with the pretrained
scale $\vb$ using $|\cos(\Delta\vb,\vb)|$, the alignment of the change with the original vector.
A value of one indicates a change along the original direction, consistent with a temperature
adjustment; $\sqrt{2/(\pi D)}$ is an approximate random-direction baseline in $D$ dimensions.
Rescaling the $\lVert\Delta\vb\rVert$-weighted mean between that baseline and one gives a
magnitude index of $0.018$--$0.241$ across module groups, indicating that the change is dominated
by reorientation rather than a uniform temperature adjustment.

\paragraph{Random-direction control.}
To test whether the learned direction matters, the control varies only the feature-embedder group
and leaves the rest of the scale stack trained. Adding a
random tensor $R$ of matched per-key Frobenius norm to the \emph{pretrained} embedder weights retains
$13.2$--$54.0\%$ over five draws, similar to reverting them outright
($8.5$--$56.6\%$); adding the same $R$ to the \emph{trained} weights leaves the result intact
($98.0$--$100.6\%$). Thus, the trained weights tolerate these perturbations, but matched-norm
random updates to the pretrained weights do not recover their benefit. The remaining trained
modules contribute to recovery in both controls.

\paragraph{DiagScale ablations.}
At learning rate $10^{-2}$, training
$\vdelta$ across all scaling modules recovers $96.3$--$137.1\%$ of the loss reduction from
jointly training both SoftScale branches. Recovery is measured against each experiment's joint
SoftScale baseline, relative to pretrained scale weights along the same inference path.
Restricting $\vdelta$ to the three feature-embedder modules (384 parameters) at the same rate
recovers $74.9$--$82.5\%$.

%% file: appendix_aggregation.tex
\section{Aggregation: implementation and additional results}
\label{app:aggregation}

\subsection{Configuration-pool construction}
\label{app:aggregation-pool}

The configuration members are drawn from TabArena's tuned TabPFN-v2.5 search space with its
one v2.5-specific entry removed, since TabPFN-3 ships a single checkpoint and publishes an empty
search space of its own. Every remaining knob is a preprocessing or inference setting that TabPFN-3
accepts natively: the softmax temperature (9 values), probability balancing (classification only),
outlier-removal threshold (3), polynomial features (2), the scaling pipeline (10 alternatives,
several of them two-stage), categorical handling (3), whether original features are appended (2), a
global transform (3), and, for regression, the target preprocessing chain (7).

The configurations are indexed $c=0,\ldots,95$.
Configuration~0 uses the default settings, so the complete $K = 96$
pool contains a default member $p_0$; the smaller pools are drawn independently and do not use it
as a forced anchor. At each pool size, the $K$ sweeps average performance over independent random
subsets: 8 draws at $K=1$ and 8, 4 at 16 and 32, 2 at 64, and the complete pool at $K=96$.
Within each test split, we average errors over subset draws and any validation subsampling
seeds before computing the log error ratio to the separately evaluated default baseline $p_{\rm base}$.

Each member is evaluated at $V = 8$ native views on the test partition and at $V = 2$ on the
validation partition. The complete pool therefore uses $96 \times 8 = 768$ test-side view
evaluations, versus 32 when using a single configuration with $V=32$; these settings are not
compute matched.
Appendix~\ref{app:runtime-accounting} reports measured costs.

\subsection{Reducers and validation protocol}
\label{app:aggregation-reducers}

For frozen-backbone aggregation, out-of-fold predictions come from cross-validation over the
cell's outer-training partition, with up to eight folds. Classification folds are stratified;
the fold count is reduced when a rare class has fewer than eight training examples. Of the
816 cells, 786 use eight folds, 20 use five, and 10 use six.

Uniform averaging assigns equal weights to all members, while best-single selects the member
with the lowest out-of-fold loss. Greedy selection \citep{caruana2004ensemble} adds the member
that minimizes the updated ensemble's out-of-fold loss, with replacement, for $T=40$ rounds.
A member selected $n_c$ times receives weight $n_c/T$.

Writing $\widehat{L}$ for the loss on the out-of-fold predictions, the reducers are
\begin{equation}
\label{eq:reducers}
\begin{aligned}
\mathcal{R}_{\mathrm{unif}}(P_{\mathcal{I}}) &= \tfrac{1}{K}\textstyle\sum_{c \in \mathcal{I}} p_c, \\
\mathcal{R}_{\mathrm{best}}(P_{\mathcal{I}}) &= p_{\hat{c}}, & \hat{c} &= \argmin_{c \in \mathcal{I}}\; \widehat{L}(p_c), \\
\mathcal{R}_{\mathrm{ges}}(P_{\mathcal{I}}) &= \tfrac{1}{T}\textstyle\sum_{t=1}^{T} p_{c_t}, & c_t &=
  \argmin_{c \in \mathcal{I}}\; \widehat{L}\Big(\tfrac{1}{t}\big(p_c + \textstyle\sum_{s<t} p_{c_s}\big)\Big).
\end{aligned}
\end{equation}
Each training row's out-of-fold prediction uses only the other folds as context. Selection uses
log loss for classification and RMSE for regression; for binary tasks, this differs from
TabArena's test metric, ROC AUC.

If a configuration fails during either validation or test prediction, its predictions on both
partitions are replaced by configuration~0. This occurs for 11 configurations in each of nine
Amazon\_employee\_access cells (99 member--cell replacements), in both the frozen and composed
pools. Thus 96 denotes configuration slots; some cells contain repeated default predictions.
The composition's additional adapter-fitting step is described in Appendix~\ref{app:composition}.

Algorithm~\ref{alg:aggregation} makes the order of prediction generation and selection explicit.
$\operatorname{Predict}_{\bar\theta,c,V}(X;D')$ fits the preprocessing and context for configuration
$c$ on $D'$ and predicts $X$ with $V$ native views, without changing $\bar\theta$.
Here $\bar\theta$ is the pretrained state for frozen aggregation. The default failure is fatal;
only nondefault members can fall back to it. Classification outputs are aligned to a common
class order before scoring and averaging. The selector uses all training-side rows unless the
label-budget analysis restricts $\Lambda$. Exact score ties are resolved by the first member in
pool order. All 40 selections are retained; the procedure does not stop when its loss plateaus.
Uniform and best-single use the same member predictions with the alternative weights in
Equation~(\ref{eq:reducers}).

\input{algorithms/aggregation}

\subsection{Aggregation results}
\label{app:aggregation-scaling}

\begin{figure}[!htb]
\centering
\includegraphics[width=\linewidth]{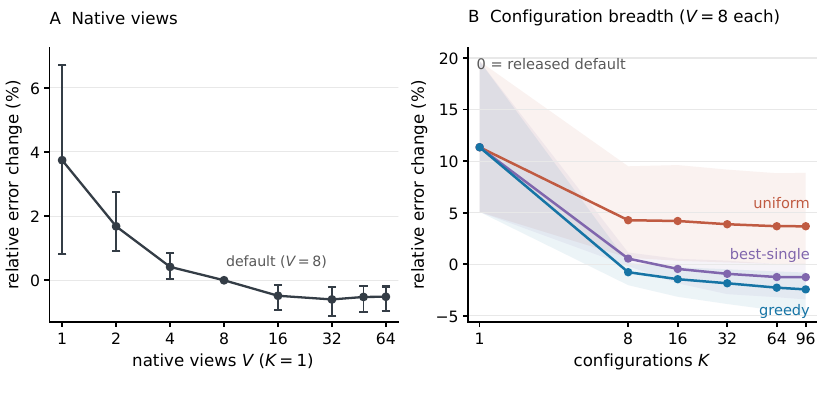}
\caption{Complete aggregation curves, including native views below the released default and
random single configurations. The three reducers coincide at $K=1$. Both panels use the same
paired estimator and transformed 95\% dataset-cluster bootstrap intervals as
Figure~\ref{fig:agg-canonical}, with independent vertical ranges and all intervals visible.
The additional $V=48$ point is included between the labeled 32- and 64-view ticks.}
\label{fig:agg-full}
\end{figure}

Figure~\ref{fig:agg-full} and Table~\ref{tab:agg-canonical} extend Figure~\ref{fig:agg-canonical}
to every evaluated view count and pool size. Let $\bar\ell$ be the mean log error ratio against
the default baseline $p_{\rm base}$ at $V{=}8$, averaged within datasets and then over 51 equally weighted
datasets. Points and the limits of the 10{,}000-resample dataset-cluster bootstrap are both
transformed with $100(\exp(\cdot)-1)$ to report percentage error changes. Panel A varies native
views of the default configuration; panel B draws configuration pools at $V{=}8$, averaging over independent
subsets. Its $K{=}1$ point therefore summarizes randomly sampled single members.

Native-view error changes are $-0.60\%$, $-0.52\%$, and $-0.51\%$ at $V=32,48,64$,
respectively, suggesting a plateau. A randomly sampled configuration increases error by
$11.35\%$ on average (95\% CI: $[5.07\%,19.69\%]$). With greedy selection, the mean point
estimate improves through $K=96$, the largest pool evaluated.
Using 32 native views and greedy selection over 96 configurations yields $+21$ and $+67$ Elo,
respectively.
Each Elo fit adds one candidate to the fixed 68 references and measures $\Delta$ Elo against
the published TabPFN-3 default within that fit; the other settings are reported in paired error.

\begin{table}[!htb]
\centering\small
\setlength{\tabcolsep}{5pt}
\caption{Every arm of Figure~\ref{fig:agg-full}, full protocol (816 cells), against the
default baseline $p_{\rm base}$ at $V{=}8$. Error is the percentage change $100(\exp(\bar\ell)-1)$ over
51 equally weighted datasets with a transformed 95\% dataset-cluster bootstrap interval;
negative is better. Configuration rows average over independent random subsets.
\emph{Wins} counts datasets with reduced error and is reported for native-view settings and
the complete configuration pool; dashes indicate counts not reported. \emph{Inner uniform}
averages native views within one configuration.}
\label{tab:agg-canonical}
\begin{tabular}{@{}rrlll@{}}
\toprule
$K$ & $V$ & \textbf{Reducer} & \textbf{$\Delta$ error [95\% CI]} & \textbf{Wins} \\
\midrule

1 &  1 & inner uniform & $+3.74\%$ $[+0.81\%, +6.70\%]$ & 4/51 \\
1 &  2 & inner uniform & $+1.68\%$ $[+0.90\%, +2.75\%]$ & 9/51 \\
1 &  4 & inner uniform & $+0.42\%$ $[+0.05\%, +0.84\%]$ & 18/51 \\
1 &  8 & inner uniform & \multicolumn{2}{c}{--- (baseline)} \\
1 & 16 & inner uniform & $-0.48\%$ $[-0.93\%, -0.14\%]$ & 33/51 \\
1 & 32 & inner uniform & $-0.60\%$ $[-1.12\%, -0.20\%]$ & 42/51 \\
1 & 48 & inner uniform & $-0.52\%$ $[-1.00\%, -0.17\%]$ & 40/51 \\
1 & 64 & inner uniform & $-0.51\%$ $[-0.96\%, -0.19\%]$ & 40/51 \\
\midrule
1 & 8 & mean single member & $+11.35\%$ $[+5.07\%, +19.69\%]$ & --- \\
8 & 8 & uniform & $+4.27\%$ $[+0.64\%, +9.50\%]$ & --- \\
16 & 8 & uniform & $+4.19\%$ $[+0.33\%, +9.62\%]$ & --- \\
32 & 8 & uniform & $+3.88\%$ $[+0.18\%, +9.18\%]$ & --- \\
64 & 8 & uniform & $+3.69\%$ $[+0.03\%, +8.83\%]$ & --- \\
96 & 8 & uniform & $+3.68\%$ $[+0.02\%, +8.86\%]$ & 14/51 \\
8 & 8 & best-single & $+0.55\%$ $[-0.18\%, +1.14\%]$ & --- \\
16 & 8 & best-single & $-0.46\%$ $[-1.85\%, +0.53\%]$ & --- \\
32 & 8 & best-single & $-0.93\%$ $[-2.90\%, +0.35\%]$ & --- \\
64 & 8 & best-single & $-1.24\%$ $[-3.20\%, +0.02\%]$ & --- \\
96 & 8 & best-single & $-1.25\%$ $[-3.41\%, +0.07\%]$ & 29/51 \\
8 & 8 & greedy & $-0.77\%$ $[-2.03\%, +0.12\%]$ & --- \\
16 & 8 & greedy & $-1.45\%$ $[-3.16\%, -0.29\%]$ & --- \\
32 & 8 & greedy & $-1.85\%$ $[-3.86\%, -0.49\%]$ & --- \\
64 & 8 & greedy & $-2.27\%$ $[-4.50\%, -0.70\%]$ & --- \\
\textbf{96} & 8 & \textbf{greedy} & \textbf{$-2.44\%$} $[-4.75\%, -0.80\%]$ & \textbf{42/51} \\
\bottomrule
\end{tabular}
\end{table}

\paragraph{Greedy selection against best-single.}
On the same $K = 96$ predictions, greedy selection lowers error by $1.20\%$ $[0.57\%, 2.03\%]$
relative to best-single and is better on 42 of 51 datasets.

\paragraph{Results by problem type.}
Table~\ref{tab:agg-task-type} summarizes the three reducers' results by task type.

\begin{table}[!htb]
\centering
\small
\caption{Relative error change against the default at $K=96$, by task type; negative is better.
Wins count datasets with reduced error.}
\label{tab:agg-task-type}
\begin{tabular}{@{}llrrrr@{}}
\toprule
\textbf{Type} & \textbf{Metric} & \textbf{Reducer} & \textbf{Median} & \textbf{Worst} & \textbf{Wins} \\
\midrule
binary (30)     & ROC AUC  & uniform     & $+0.30\%$ & $+5.29\%$   &  9/30 \\
                &          & best-single & $+0.05\%$ & $+4.20\%$   & 13/30 \\
                &          & greedy      & $-0.31\%$ & $+1.02\%$   & \textbf{22/30} \\
multiclass (8)  & log loss & uniform     & $+5.80\%$ & $+164.30\%$ &  1/8 \\
                &          & best-single & $-0.85\%$ & $+0.35\%$   &  7/8 \\
                &          & greedy      & $-1.21\%$ & $-0.32\%$   & \textbf{8/8} \\
regression (13) & RMSE     & uniform     & $+0.57\%$ & $+2.20\%$   &  4/13 \\
                &          & best-single & $-0.12\%$ & $+0.58\%$   &  9/13 \\
                &          & greedy      & $-0.43\%$ & $+0.21\%$   & \textbf{12/13} \\
\bottomrule
\end{tabular}
\end{table}

\subsection{Dependence on validation data}
\label{app:aggregation-validation}

\begin{figure}[!htb]
\centering
\includegraphics[width=\linewidth]{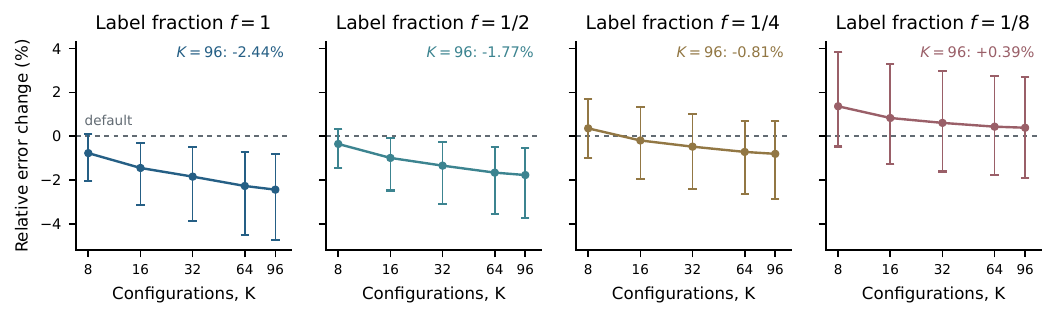}
\caption{Greedy aggregation with different fractions of training-side selector labels.
All panels share the same axes and fixed member predictions. Points use the dataset-weighted
mean log-error ratio against the released default; bars show transformed 95\% dataset-bootstrap
intervals. Pool breadth improves each curve's point estimates, while fewer selector labels
reduce the gains. Intervals exclude zero for $f\in\{1/2,1\}$ and $K\ge16$;
all others include zero.}
\label{fig:agg-label-budget}
\end{figure}

Figure~\ref{fig:agg-label-budget} and Table~\ref{tab:agg-kf} give the full $K \times f$ surface for greedy selection, where $f$ is the
fraction of out-of-fold labels the reducer may use. Member predictions are fixed, and
within a cell the configuration subset drawn at each $(K, \text{draw})$ is held fixed across every
$f$ and every label seed, so a difference along a row is a difference in the reducer's label access
and not in which members it saw. Label subsets are nested across budgets for a fixed seed and
sampled inside each (fold, class) group, using eight seeds for $f < 1$. Errors are averaged over
seeds and configuration draws within each cell before computing log error ratios; these are
averaged within datasets, with equal weight across the 51 datasets.

\begin{table}[!htb]
\centering
\small
\caption{Greedy selection: relative error change against the default baseline $p_{\rm base}$ at $V{=}8$, by
pool size and out-of-fold label fraction; negative is better. The last row is the change from
$K = 8$ to $K = 96$ at that budget, in percentage points. The $K = 1$ row is the mean over randomly
sampled single-member pools, not $p_0$; the reducer has nothing to choose between there, so it does
not vary with $f$.}
\label{tab:agg-kf}
\begin{tabular}{@{}lrrrr@{}}
\toprule
$K$ & $f = 0.125$ & $f = 0.25$ & $f = 0.5$ & $f = 1.0$ \\
\midrule
1   & $+11.35\%$ & $+11.35\%$ & $+11.35\%$ & $+11.35\%$ \\
8   & $+1.37\%$  & $+0.36\%$  & $-0.35\%$  & $-0.77\%$ \\
16  & $+0.83\%$  & $-0.19\%$  & $-0.99\%$  & $-1.45\%$ \\
32  & $+0.61\%$  & $-0.48\%$  & $-1.34\%$  & $-1.85\%$ \\
64  & $+0.44\%$  & $-0.72\%$  & $-1.66\%$  & $-2.27\%$ \\
96  & $+0.39\%$  & $-0.81\%$  & $-1.77\%$  & $-2.44\%$ \\
\midrule
$8 \to 96$ & $-0.98$ & $-1.17$ & $-1.42$ & $-1.67$ \\
\bottomrule
\end{tabular}
\end{table}

At every label budget, larger pools improve the point estimates, but with only one eighth of
the labels, point estimates remain above the default for all evaluated pool sizes. Because member predictions
and configuration subsets are fixed across budgets, these differences reflect the reducer's
access to validation labels.

%% file: algorithms/aggregation.tex
\begin{algorithm}[!htb]
\caption{Configuration predictions and greedy aggregation}
\label{alg:aggregation}
\small
\begin{algorithmic}[1]
\Require Outer-training table $D$, unlabeled queries $X_Q$, fixed model state $\bar\theta$
\Require Configuration universe $\mathcal C$ containing default $0$, retained pool $\mathcal I\subseteq\mathcal C$
\Require Selector rows $\Lambda\subseteq\{1,\ldots,|D|\}$, loss $L$, selection rounds $T=40$
\Ensure Mixture weights $w$ and query predictions $\hat P$
\State Partition $D$ into $F\leq8$ inner folds; stratify classification and reduce $F$ for rare classes
\For{each configuration $c\in\mathcal C$, with default $0$ first}
  \For{each inner fold $I_f$}
    \State $P_c[I_f]\gets\operatorname{Predict}_{\bar\theta,c,2}(X_{I_f};D_{-I_f})$
  \EndFor
\EndFor
\For{each configuration $c\in\mathcal C$, with default $0$ first}
  \State $U_c\gets\operatorname{Predict}_{\bar\theta,c,8}(X_Q;D)$
\EndFor
\State If member $c$ fails at either stage, replace both $(P_c,U_c)$ by $(P_0,U_0)$
\State $A\gets0$ (selector prediction sum); $n_c\gets0$ for $c\in\mathcal I$
\For{$t=1,\ldots,T$}
  \State $c_t\gets\arg\min_{c\in\mathcal I}L\!\left(y_\Lambda,(A+P_c[\Lambda])/t\right)$
  \State $A\gets A+P_{c_t}[\Lambda]$; $n_{c_t}\gets n_{c_t}+1$
\EndFor
\State $w_c\gets n_c/T$; $\hat P\gets\sum_{c\in\mathcal I}w_cU_c$
\State \Return $(w,\hat P)$
\end{algorithmic}
\end{algorithm}

%% file: appendix_context.tex
\section{Context construction: implementation and additional results}
\label{app:context}

\subsection{Shared setup and terminology}
\label{app:context-setup}

All studies keep the backbone frozen and change its conditioning context. References and view
counts are specified per experiment. The main column comparison uses eight views and matches
the default's per-view column counts. Large-row studies use a single view and compare with
full-context inference (\emph{feed-all}) at the same source-pool size $N$ within memory, or at 4M rows beyond capacity.
\emph{Model-aware} denotes attention-guided row retrieval.
Expansion compares with the same context without added rows or columns.

Benchmark experiments use their stated task metric; large-row experiments use log loss,
including binary Higgs and COMET. Relative error reductions are
$100\times(\text{reference}-\text{arm})/\text{reference}$, so \emph{positive is better}.
Figure~\ref{fig:row-curation} uses matched random contexts as the reference in panel A and
absolute log loss in panels B/C. Absolute errors are compared within a dataset.

\subsection{Row curation and retrieval}
\label{app:row-curation}

\paragraph{The model-aware pipeline.} We adapt the released retrieval pipeline of
\citet{zhang2025limix} to frozen TabPFN-3; Algorithm~\ref{alg:model-aware-retrieval} gives the
evaluated procedure. Retrieval uses the last ICL layer's query-to-context softmax attention,
averaged over heads, with retrieval budget $r=\min(500,N)$. A binary indicator represents
each query by the source rows it retrieves. Queries with similar retrieval
sets are clustered, and each cluster receives the unique union of its selected rows.

We use $G=\min(n_Q,\max(4,\min(30,\lfloor n_Q/2000\rfloor)))$ query clusters, with four KMeans
initializations and random seed 0. This gives 25 initial clusters for the 50k-query large-row studies.
Scoring uses one native view; per-cluster deployment uses one view in the large-row sweep and
one or eight in the TabArena comparison. Both scoring and clustering access the
unlabeled query batch, so the constructed contexts depend on which queries are available together;
query labels are used only for evaluation. Classification probabilities are aligned to the source
pool's class order, with zero probability for classes missing from a selected context.

\input{algorithms/model_aware_retrieval}

\paragraph{Reaching past the memory wall.} Chunked TabPFN uses tiled attention to support longer
contexts without additional training \citep{sergazinov2025chunked}.
In our setup, feed-all becomes infeasible at 4.0--4.5M rows on an 80\,GB device.
Beyond this limit, we shuffle the pool and score attention separately in 2M-row blocks.
Each query retrieves a proportional quota $r|B_b|/N$ from each block, rounded to total $r$
(Algorithm~\ref{alg:model-aware-retrieval}). Selection uses within-block ranks: the block
softmaxes have different denominators and do not define a global top-$r$ ranking.
Unsharded scoring uses the whole pool as one block.
At 4M rows, two-block scoring increases retrieval log loss relative to unsharded scoring by
$0.01\%$ on Higgs, $0.68\%$ on US\_Accidents, and $57.1\%$ on COMET\_MC,
averaged over five paired seeds. All five COMET\_MC seeds worsen; the retrieval failure also
occurs without sharding.

\paragraph{Union capacity.} On Higgs, query clusters whose retrieved union exceeds $C=4$M rows
are recursively split into balanced halves using their retrieval indicators, until every union
fits. All retrieved rows are retained. The 6M--10M runs use 26--27 final clusters, below the
configured maximum of 64; splitting uses no query labels or prediction losses. Other datasets and the capped
Higgs control keep only the $C$ rows with the largest summed retained attention when a union
exceeds capacity. For sharded scoring, these values are block-local and are not calibrated
across blocks. At 7.68M rows on US\_Accidents, this cap removes 14.8--16.9\% of one cluster's
union in each of the five seeds. Table~\ref{tab:higgs-union-capacity} compares the two policies.

\paragraph{The geometric coreset.} This comparison clusters query rows in feature space after
training-mean imputation and training-fitted standardization. Each cluster is predicted using
the $100{,}000$ source rows nearest its centroid, with a separate frozen fit per cluster.

\begin{figure}[!htbp]
\centering
\includegraphics[width=\linewidth]{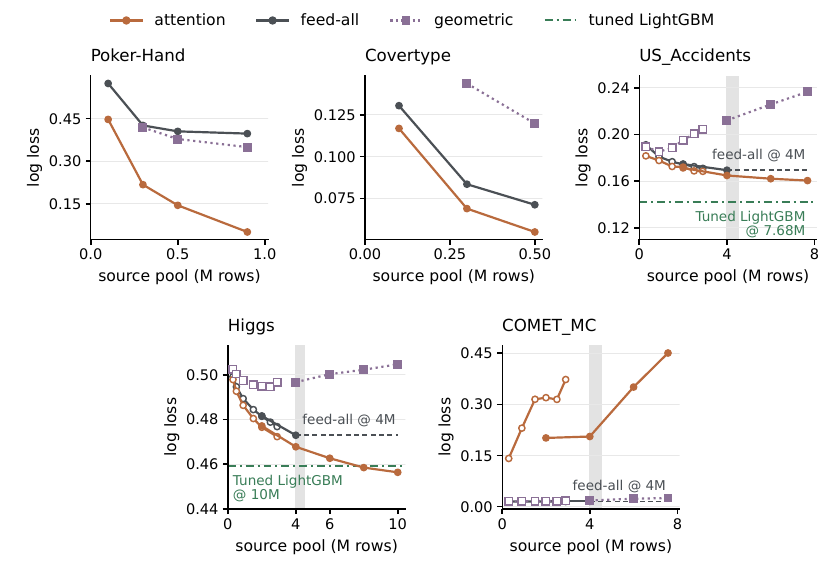}
\caption{Complete source-pool sweeps with single-view TabPFN-3 on five large tables, including the geometric construction
omitted from the main figure. Points are the five-seed mean log losses; each panel has an
independent linear vertical axis. For the three largest tables, open and filled markers distinguish
the 3M-subset and full-table splits; lines connect only points from the same split.
Feed-all curves contain only measured values, ending at 4M where
the pool exceeds capacity. Beyond 4M, dashed lines hold that reference fixed and gray bands mark
the 4.0--4.5M capacity boundary. Green lines on US\_Accidents and Higgs are tuned LightGBM
endpoint references (Table~\ref{tab:row-lightgbm}).}
\label{fig:row-curation-full}
\end{figure}

The large-row sweeps use five seeds per setting. On Higgs, US\_Accidents, and COMET\_MC,
the sweep through 2.9M rows uses a 3M-row source subset, while the 2M-and-larger sweep uses the
full table. Each source has its own seed-specific 50k-row test set; the two sweeps are plotted
separately. Table~\ref{tab:row-construction} gives selected sizes from
Figure~\ref{fig:row-curation-full}, together with each dataset's complete source pool. Within the memory
limit both constructions are measured against feed-all at the same $N$; beyond it, where no such run
exists, the reference is feed-all at 4M from the full-table sweep.

\begin{table}[h]
\centering
\small
\caption{Row construction on large tables, frozen TabPFN-3, single view. Relative log-loss reduction
against the full-context reference for the regime, in per cent; positive is better. Within memory
the reference is feed-all at the same $N$; beyond the memory wall no such run exists, so it is
feed-all at 4M. A dash marks a source pool that never reaches the wall. Entries summarize five
seeds. The 2.9M results use the 3M-subset splits; beyond-capacity results and their 4M references
use the full-table splits.}
\label{tab:row-construction}
\begin{tabular}{@{}lrrrrl@{}}
\toprule
 & \textbf{Total} & \multicolumn{3}{c}{\textbf{Within memory}} & \textbf{Beyond wall} \\
\cmidrule(lr){3-5}
\textbf{Dataset} & $N$ & \textbf{Eval.\ }$N$ & \textbf{Geo.} & \textbf{Model-aware} & \textbf{Model-aware} \\
\midrule
Poker-Hand   & 1.03M & 900k & $+12.0$ & $\mathbf{+87.2}$ & --- \\
Covertype    & 0.58M & 500k & $-67.9$ & $\mathbf{+22.9}$ & --- \\
US\_Accidents & 7.73M & 2.9M & $-19.8$ & $+1.5$ & $\mathbf{+5.2}$ at 7.68M \\
Higgs        & 11M   & 2.9M & $-4.2$  & $+0.9$ & $+2.2$ at 6M, $\mathbf{+3.5}$ at 10M \\
COMET\_MC    & 7.62M & 2.9M & $-11.3$ & $-2375$ & $-2808$ at 7.57M \\
\bottomrule
\end{tabular}
\end{table}

\paragraph{Higgs capacity check.}
The same splitting rule is used at 6M, 8M, and 10M rows, with no truncation in any run.
Mean log loss decreases through 10M rows, whereas the capped control rises after 6M
(Table~\ref{tab:higgs-union-capacity}). Splitting also changes the
query partition, so this comparison does not isolate truncation from grouping.
Seed 0 informed development; excluding it gives means $0.46258$, $0.45874$, and $0.45653$
over the other four seeds.

\begin{table}[h]
\centering
\small
\caption{Higgs union-capacity comparison. Log loss is mean $\pm$ standard deviation over five
seeds. Time includes attention scoring and is relative to feed-all at 4M rows.
Both variants keep the retrieval budget and scoring shards fixed.}
\label{tab:higgs-union-capacity}
\begin{tabular}{@{}lrrrr@{}}
\toprule
\textbf{Pool} & \textbf{Capped union} & \textbf{Recursive splitting} & \multicolumn{2}{c}{\textbf{Time ($\times$)}} \\
\cmidrule(lr){4-5}
 & & & \textbf{Capped} & \textbf{Split} \\
\midrule
6M & $0.46279\pm0.00316$ & $0.46266\pm0.00343$ & $1.95$ & $2.32$ \\
8M & $0.46868\pm0.00283$ & $0.45848\pm0.00328$ & $2.20$ & $3.16$ \\
10M & $0.48530\pm0.00362$ & $0.45637\pm0.00279$ & $2.47$ & $3.95$ \\
\bottomrule
\end{tabular}
\end{table}

\paragraph{Validation-tuned LightGBM reference.}
\label{app:row-lightgbm}
The reference uses the same deterministic per-seed permutation and 50{,}000 held-out test rows as
the full-pool row study. For each of five seeds, it selects among learning rates
$\{0.02,0.05,0.1\}$ and leaf counts $\{63,255\}$ by validation log loss.
Every candidate allows at most 100{,}000 boosting rounds, with early stopping after 300 rounds
without validation improvement. Table~\ref{tab:row-lightgbm} reports the selected configurations;
all selected models stop before the cap. Test loss does not select the configuration.

For US\_Accidents, the 7{,}678{,}394-row pool is split into 7{,}524{,}827 fitting rows and
153{,}567 validation rows. For Higgs, the reference fits on the 10{,}000{,}000-row pool
and validates on the next 200{,}000 rows of the remaining training data, outside that pool.
Categorical columns with at most 255 distinct training values are declared categorical; higher
cardinality columns retain their numeric codes. The runs use 24 CPU threads.
The horizontal lines are five-seed endpoint means, $0.14182$ for US\_Accidents and $0.45909$
for Higgs; they do not represent models tuned separately at every plotted source-pool size.

\begin{table}[h]
\centering
\small
\caption{Validation-tuned LightGBM endpoints, five seeds. All selected models use 255 leaves.
US\_Accidents fits 7{,}524{,}827 rows with 153{,}567 validation rows; Higgs fits 10M rows with
200k additional validation rows. ``Rounds'' is the selected best iteration under a 100k cap.}
\label{tab:row-lightgbm}
\begin{tabular}{@{}lrrrrr@{}}
\toprule
\textbf{Dataset} & \textbf{Seed} & \textbf{LR} & \textbf{Rounds} & \textbf{Val. loss} & \textbf{Test loss} \\
\midrule
US\_Accidents & 0 & 0.02 & 7{,}972 & 0.14011 & 0.13994 \\
              & 1 & 0.02 & 8{,}093 & 0.14328 & 0.14279 \\
              & 2 & 0.02 & 8{,}056 & 0.13990 & 0.14081 \\
              & 3 & 0.02 & 8{,}411 & 0.13998 & 0.14279 \\
              & 4 & 0.02 & 7{,}479 & 0.14216 & 0.14276 \\
\midrule
Higgs & 0 & 0.02 & 72{,}769 & 0.45890 & 0.45808 \\
      & 1 & 0.02 & 87{,}121 & 0.45929 & 0.46381 \\
      & 2 & 0.05 & 32{,}073 & 0.45931 & 0.45968 \\
      & 3 & 0.02 & 76{,}666 & 0.45639 & 0.45530 \\
      & 4 & 0.02 & 82{,}420 & 0.45694 & 0.45859 \\
\bottomrule
\end{tabular}
\end{table}

\paragraph{The benchmark-scale sweep.} Table~\ref{tab:row-tabarena} gives all 15 datasets,
each evaluated with ten repeats of three-fold cross-validation, including datasets assigned
only three repeats in the main TabArena protocol.
The mean attention mass retained by the rows selected for each query ranges from
$100\,\%$ on blood-transfusion to $14\,\%$ on diamonds.
Under the eight-view default, the corrected intervals exclude zero on four datasets,
two positive and two negative; the sign agrees
between the single-view and eight-view arms on every dataset that resolves under both.

For each split $j$, we compute the paired relative error reduction
$d_j=100(1-e_{\mathrm{retrieval},j}/e_{\mathrm{full},j})$ and report its mean $\bar d$.
To account approximately for overlapping training sets, we use the corrected resampled
$t$ interval of \citet{nadeau1999inference}:
\begin{equation}
\bar d \;\pm\; t_{29,0.975}\,s_d
\sqrt{\frac{1}{30}+\bar q},
\qquad
\bar q=\frac{1}{30}\sum_{j=1}^{30}\frac{n_{\mathrm{test},j}}{n_{\mathrm{train},j}},
\label{eq:row-corrected-interval}
\end{equation}
where $s_d$ is the sample standard deviation of the 30 reductions and $\bar q\approx1/2$.
These are nominal dataset-wise intervals without adjustment for multiple comparisons.

\begin{table}[h]
\centering
\small
\setlength{\tabcolsep}{4pt}
\caption{Model-aware row construction at TabArena scale, all 15 datasets, 30 splits per dataset.
$n$ is the maximum number of training rows across splits, $k$ the number of query clusters,
\emph{attn.} the mean attention mass retained by
the rows selected for each query, and \emph{union} the median per-cluster context size. The last two columns
give mean paired relative error reductions at one view and at the released eight-view default,
in per cent, with approximate $95\,\%$ intervals from Equation~(\ref{eq:row-corrected-interval});
positive is better. Bold marks intervals excluding zero.}
\label{tab:row-tabarena}
\begin{tabular}{@{}lrrrrrr@{}}
\toprule
\textbf{Dataset} & $n$ & $k$ & \textbf{attn.\ \%} & \textbf{union} & \textbf{$\Delta$ 1 view} & \textbf{$\Delta$ 8 views} \\
\midrule
blood-transfusion-service-center         & 499 & 4 & 100 & 499 & $+0.00$ $\pm0.00$ & $+0.00$ $\pm0.00$ \\
diabetes                                 & 512 & 4 & 99 & 512 & $+0.00$ $\pm0.85$ & $+0.04$ $\pm0.45$ \\
maternal\_health\_risk                   & 676 & 4 & 91 & 634 & $+2.98$ $\pm5.72$ & $+0.77$ $\pm2.43$ \\
concrete\_compressive\_strength          & 687 & 4 & 93 & 684 & $-0.57$ $\pm1.96$ & $-0.12$ $\pm0.80$ \\
airfoil\_self\_noise                     & 1{,}002 & 4 & 83 & 972 & $+0.63$ $\pm3.58$ & $-0.29$ $\pm1.23$ \\
students\_dropout\_and\_academic\_success & 2{,}950 & 4 & 50 & 1{,}814 & $-0.25$ $\pm1.24$ & $-0.19$ $\pm0.62$ \\
churn                                    & 3{,}334 & 4 & 46 & 2{,}198 & $-2.82$ $\pm7.52$ & $\mathbf{-4.29}$ $\pm3.89$ \\
wine\_quality                            & 4{,}332 & 4 & 41 & 2{,}726 & $\mathbf{-9.02}$ $\pm1.69$ & $\mathbf{-8.94}$ $\pm1.66$ \\
Bank\_Customer\_Churn                    & 6{,}667 & 4 & 29 & 3{,}672 & $\mathbf{-0.90}$ $\pm0.80$ & $-0.54$ $\pm0.61$ \\
heloc                                    & 6{,}973 & 4 & 26 & 4{,}146 & $-0.55$ $\pm0.68$ & $-0.53$ $\pm0.54$ \\
houses                                   & 13{,}760 & 4 & 32 & 9{,}878 & $+0.04$ $\pm0.80$ & $+0.19$ $\pm0.38$ \\
credit\_card\_clients\_default           & 20{,}000 & 5 & 21 & 7{,}784 & $-0.18$ $\pm0.40$ & $-0.15$ $\pm0.30$ \\
bank-marketing                           & 30{,}141 & 7 & 16 & 10{,}105 & $-0.10$ $\pm0.46$ & $-0.09$ $\pm0.42$ \\
diamonds                                 & 35{,}960 & 8 & 14 & 9{,}446 & $\mathbf{+0.67}$ $\pm0.60$ & $\mathbf{+0.61}$ $\pm0.36$ \\
SDSS17                                   & 52{,}036 & 13 & 22 & 8{,}840 & $\mathbf{+9.35}$ $\pm3.85$ & $\mathbf{+5.36}$ $\pm1.47$ \\
\bottomrule
\end{tabular}
\end{table}

\subsection{Matched-row controls and the COMET failure}
\label{app:comet-diagnostic}

The model-aware pipeline changes several things at once relative to feed-all: it shrinks the context,
partitions the queries, and refits once per partition. To separate those from the choice of rows, we
hold attention scoring, query clustering, per-cluster context sizes, refitting, and prediction
stitching fixed. The control replaces only the \emph{identities} of selected rows with a uniform
random draw of the same size. The controls use
three paired seeds, except COMET, which uses five. All arms use matched test splits.

Equivalently, after constructing and capping $U_g$ in Algorithm~\ref{alg:model-aware-retrieval},
the identity control replaces it by a uniform sample without replacement of $|U_g|$ source rows.
The attention-derived query clusters and all subsequent prediction steps are shared.

For Figure~\ref{fig:row-curation}A and Table~\ref{tab:identity-control}, define the paired reduction
$r_s=100(1-e_{s,\mathrm{attention}}/e_{s,\mathrm{random}})$ for each seed $s$.
We report its arithmetic mean with interval
$\bar r\pm t_{0.975,n-1}\,\mathrm{sd}(r)/\sqrt n$.
The main panel shows Covertype, Poker, and both Higgs settings; the COMET control is reported in Table~\ref{tab:comet-diagnostic}.

\begin{table}[h]
\centering
\small
\caption{Matched-row identity control. Log loss, mean over three seeds. The final column is the relative
change from the random-identity arm to the attention arm; positive means the attention-selected rows
are better. This is the contrast plotted in Figure~\ref{fig:row-curation}A.
All four 95\% intervals exclude zero.}
\label{tab:identity-control}
\begin{tabular}{@{}lrrrrl@{}}
\toprule
\textbf{Dataset} & $N$ & \textbf{Feed-all} & \textbf{Random ids} & \textbf{Attention} & \textbf{Attention vs.\ random} \\
\midrule
Covertype & 500k & $0.0728$ & $0.1613$ & $0.0558$ & $+65.3\%$ $[+60.0, +70.6]$ \\
Poker-Hand & 900k & $0.3928$ & $0.5272$ & $0.0469$ & $+91.1\%$ $[+89.1, +93.1]$ \\
Higgs & 900k & $0.4885$ & $0.5018$ & $0.4859$ & $+3.2\%$ $[+2.8, +3.6]$ \\
Higgs & 2.9M & $0.4759$ & $0.4911$ & $0.4717$ & $+3.9\%$ $[+3.5, +4.3]$ \\
\bottomrule
\end{tabular}
\end{table}

The random-identity arm is worse than feed-all in all four settings, while attention-selected
rows improve on it. Replacing row identities therefore reverses the observed benefit conditional
on the fixed clusters, context sizes, and refit pipeline.

\paragraph{COMET\_MC.}
On COMET\_MC, the matched random control holds each cluster's context size fixed at the value
produced by attention selection. For each cluster, we examine its query count, the number of
retrieval draws and distinct rows, and the class composition of its selected context relative
to the matched random context and source pool (Table~\ref{tab:comet-diagnostic}).

\begin{table}[h]
\centering
\small
\caption{COMET\_MC at $N = 900$k, five seeds, source pool $98.3\% / 1.7\%$. Every constructed arm
refits on the same $25$ clusters with a median context of $977$ rows; only the contents differ. The
rare-class share is the median over $125$ cluster observations.}
\label{tab:comet-diagnostic}
\begin{tabular}{@{}lrrrr@{}}
\toprule
\textbf{Arm} & \textbf{Log loss} & \textbf{vs.\ feed-all} & \textbf{AUC} & \textbf{Rare-class share} \\
\midrule
Feed-all ($N = 900$k)          & $0.0151$ & ---        & $0.998$ & $0.017$ \\
Attention union                & $0.2302$ & $-1425\%$  & $0.950$ & $\mathbf{0.403}$ \\
Random identities, same size   & $0.0158$ & $-4.9\%$   & $0.998$ & $0.017$ \\
Class-marginal repair          & $0.1270$ & $-741\%$   & $0.569$ & $0.017$ \\
\bottomrule
\end{tabular}
\end{table}

At the same context sizes and with the same clusters and refits, random selection brings log
loss within $5\%$ of feed-all, so context size alone does not explain the failure. The retrieved
contexts have a median rare-class share of $40.3\%$, compared with $1.7\%$ in the source pool.
Despite each query retrieving 500 rows, their unions contain fewer than a thousand distinct
rows at the median, compared with tens to hundreds of thousands in the other settings
(Table~\ref{tab:union-collapse}).

\begin{table}[h]
\centering
\small
\caption{Retrieval concentration within a query cluster. Query count, draws, and union size are
medians over clusters and seeds (five seeds for COMET, three for the other settings).
``Overlap'' is the ratio of the reported median draws to median union size. Poker and Higgs
statistics use the matched random-control runs, whose context sizes equal those of attention selection.}
\label{tab:union-collapse}
\begin{tabular}{@{}lrrrrr@{}}
\toprule
\textbf{Dataset} & $N$ & $|Q|$ & \textbf{Draws} & \textbf{Union} & \textbf{Overlap} \\
\midrule
COMET\_MC & 900k & $2{,}270$ & $1{,}135{,}000$ & $977$ & $\mathbf{1162\times}$ \\
Covertype & 500k & $1{,}487$ & $743{,}500$ & $40{,}794$ & $18\times$ \\
Poker-Hand & 900k & $1{,}570$ & $785{,}000$ & $73{,}731$ & $11\times$ \\
Higgs & 900k & $1{,}383$ & $691{,}500$ & $85{,}188$ & $8\times$ \\
Higgs & 2.9M & $1{,}202$ & $601{,}000$ & $154{,}159$ & $4\times$ \\
\bottomrule
\end{tabular}
\end{table}

Class-distribution diagnostics retain the source pool's class identities even when their
frequencies reverse in a retrieved union. The median total variation between the retrieved
and source class distributions is $0.386$; per-cluster error correlates with this shift
(Spearman $+0.81$). Union size and rare-class share are themselves correlated ($-0.59$).
These descriptive associations over 125 cluster observations nested within five seeds do not
separate the effects of context size and class composition.

At fixed context size, class-marginal repair lowers mean log loss from $0.2302$ to $0.1270$
but does not restore predictive performance; AUC falls to $0.569$. The repair replaces the
lowest-attention rows of the over-represented class with uniformly sampled rows of the
under-represented class, changing both row identities and class composition.

\subsection{Column curation}
\label{app:column-curation}

\paragraph{Selector and protocol.}
We use TabPFN's LightGBM feature-importance implementation, which fits
LightGBM on the training split for each dataset and seed and ranks columns by total split gain.
We compute one ranking and fix its top-ranked columns across all eight views; the native
importance-based sampler instead retains top-ranked columns and fills the remaining per-view
budget by balanced sampling. Nothing from the test split enters the ranking. The main comparison keeps the
top 200 columns on six classification tables and the top 500 on QSAR-TID-11 regression, matching
TabPFN-3's default per-view caps. All eight views receive the same selected set. These counts
refer to original columns; each view still applies its native preprocessing.
The 500-column grid covers eight wide tables with $D>1000$. Training contexts are capped at
6{,}000 rows. Each setting uses four seeds and the task metric in
Table~\ref{tab:column-grid}.

\paragraph{Arms.}
The unselected one-view arm disables feature subsampling and uses all $D$ columns. The default
eight-view arm uses the backbone's balanced column sampler. In the $F=500$ controls,
importance selection fixes the top 500 columns and random selection fixes a uniform 500-column
set. The importance ranking is reused across the selection arms of each dataset/seed split.
The one-view selected arms use all 500 columns in one view; the pinned arms give all 500 to every
one of eight views. The per-view arms restore balanced subsampling within that candidate set,
using 200 columns per classification view and 500 per regression view. Thus the pinned/per-view
contrast changes both view diversity and actual per-view feature count. The importance-versus-random
contrast holds these budgets equal within each view setting. The additional 200-column
classification setting uses the eight-view pinned importance arm.

\begin{table}[!htbp]
\centering
\small
\setlength{\tabcolsep}{5.5pt}
\caption{Column curation on seven wide tables, frozen TabPFN-3. Entries are relative error
reductions from the reference in each header; positive is better. The main selected result fixes
200 columns across all eight views for classification and 500 for QSAR-TID-11 regression,
matching the default's per-view counts. ``1 view'' uses all $D$ columns. The two rightmost columns
report the separate $F=500$ controls: fixed views each use all 500 selected columns, while
per-view subsampling uses the native cap; importance and random selection are compared with
500 columns fixed across eight views. All percentages use unrounded four-seed mean errors.
Task metrics and absolute errors appear in Table~\ref{tab:column-grid}.}
\label{tab:column-curation}
\begin{tabular}{@{}llrrrrr@{}}
\toprule
 & & & \textbf{Default} & \textbf{Selected} & \multicolumn{2}{c}{\textbf{$F=500$ controls}} \\
\cmidrule(lr){6-7}
\textbf{Dataset} & \textbf{Domain} & $D$ & \textbf{vs.\ 1 view} & \textbf{vs.\ default} & \textbf{Per view vs.\ fixed} & \textbf{Imp.\ vs.\ random} \\
\midrule
Bioresponse & molecular & $1{,}776$ & $+8.9$ & $-1.4$ & $+4.5$ & $+4.5$ \\
hiva\_agnostic & molecular & $1{,}617$ & $+12.8$ & $-4.5$ & $+4.7$ & $-2.7$ \\
QSAR-TID-11 & molecular & $1{,}024$ & $+2.0$ & $+0.5$ & $+0.0$ & $+2.5$ \\
CIFAR\_10 & image & $3{,}072$ & $+6.6$ & $-2.1$ & $+1.2$ & $+0.0$ \\
\midrule
micro-mass & mass-spec & $1{,}300$ & $-10.5$ & $+35.1$ & $-7.1$ & $+19.2$ \\
11\_Tumors & genomics & $12{,}533$ & $+12.9$ & $+29.8$ & $+3.6$ & $+14.1$ \\
amazon-reviews & text & $10{,}000$ & $-43.0$ & $+67.1$ & $-31.8$ & $+65.6$ \\
\bottomrule
\end{tabular}
\end{table}

\paragraph{The complete grid.} Table~\ref{tab:column-grid} reports nine settings on eight tables,
plus the 200-column, eight-view importance results for six classification tables.
The main regression result uses 500 columns. All entries average four seeds;
\texttt{Prostate} is shown only in the 500-column grid and excluded from the summaries because its
default reference has zero error. For every percentage in Table~\ref{tab:column-curation}, we
first average the errors across seeds, then compute $100(1-\bar e_{\rm arm}/\bar e_{\rm ref})$.
Rounding is applied only to the displayed values.

\begin{table}[h]
\centering
\footnotesize
\setlength{\tabcolsep}{1.8pt}
\caption{Complete column-curation grid: mean error over 4 seeds, in each dataset's own metric, so
values compare across a row and not down a column; lower is better. \emph{gini} denotes the
backbone's importance-based per-view sampler, which also uses LightGBM gain. The
\emph{Importance} and \emph{Random} groups both keep $F=500$ columns; \emph{pin} gives all eight
views the same set, and \emph{p.v.} restores native subsampling within it. The last column adds
importance selection with 200 columns fixed across eight views. Dashes denote unavailable
200-column results; regression retains the 500-column result in the main comparison.}
\label{tab:column-grid}
\begin{tabular}{@{}llrrrrrrrrrr@{}}
\toprule
 & & \multicolumn{3}{c}{\textbf{No selection}} & \multicolumn{3}{c}{\textbf{Importance, $F=500$}} & \multicolumn{3}{c}{\textbf{Random, $F=500$}} & \textbf{$F=200$} \\
\cmidrule(lr){3-5}\cmidrule(lr){6-8}\cmidrule(lr){9-11}\cmidrule(lr){12-12}
\textbf{Dataset} & \textbf{Metric} & \textbf{1 v.} & \textbf{8 v.} & \textbf{gini} & \textbf{1 v.} & \textbf{pin} & \textbf{p.v.} & \textbf{1 v.} & \textbf{pin} & \textbf{p.v.} & \textbf{Imp.\ pin} \\
\midrule
Bioresponse & 1$-$AUC & $0.1427$ & $0.1300$ & $0.1320$ & $0.1394$ & $0.1331$ & $0.1270$ & $0.1459$ & $0.1393$ & $0.1358$ & $0.1319$ \\
hiva\_agnostic & log loss & $0.2085$ & $0.1818$ & $0.1840$ & $0.2121$ & $0.1912$ & $0.1822$ & $0.1951$ & $0.1862$ & $0.1831$ & $0.1899$ \\
QSAR-TID-11 & RMSE & $0.7432$ & $0.7282$ & $0.7269$ & $0.7312$ & $0.7246$ & $0.7244$ & $0.7624$ & $0.7429$ & $0.7414$ & --- \\
CIFAR\_10 & log loss & $1.6399$ & $1.5310$ & $1.5531$ & $1.6099$ & $1.5523$ & $1.5330$ & $1.6267$ & $1.5529$ & $1.5440$ & $1.5633$ \\
micro-mass & log loss & $0.4759$ & $0.5257$ & $0.3358$ & $0.4579$ & $0.3478$ & $0.3726$ & $0.5579$ & $0.4307$ & $0.5137$ & $0.3409$ \\
11\_Tumors & log loss & $0.3971$ & $0.3460$ & $0.2292$ & $0.3104$ & $0.2428$ & $0.2340$ & $0.3658$ & $0.2827$ & $0.3560$ & $0.2430$ \\
amazon-reviews & log loss & $1.6620$ & $2.3764$ & $0.7896$ & $0.8936$ & $0.7859$ & $1.0362$ & $2.3523$ & $2.2850$ & $2.6496$ & $0.7820$ \\
Prostate & 1$-$AUC & $0.0156$ & $0.0000$ & $0.0039$ & $0.0195$ & $0.0039$ & $0.0000$ & $0.0000$ & $0.0000$ & $0.0000$ & --- \\
\bottomrule
\end{tabular}
\end{table}

\begin{figure}[!htb]
\centering
\includegraphics[width=\linewidth]{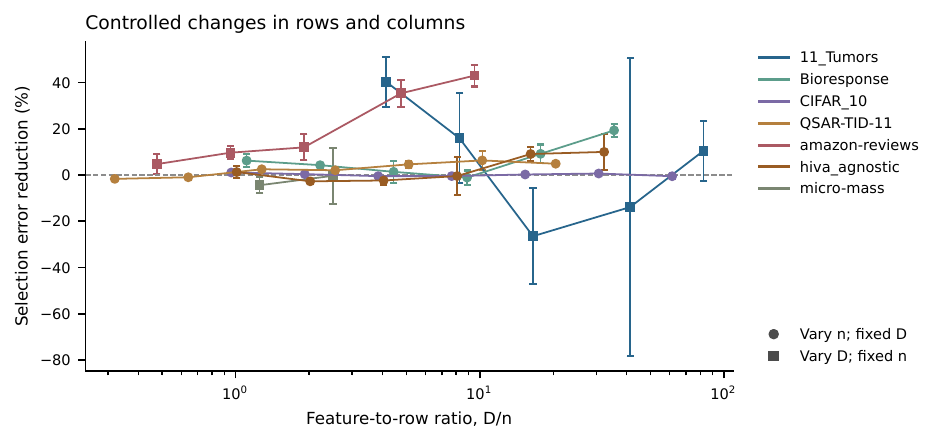}
\caption{Column-curation controls at a fixed 200-column selection budget with single-view
TabPFN-3, separate from the eight-view comparison in Table~\ref{tab:column-curation}.
Circles vary context rows at fixed width; squares vary width at fixed context size.
The LightGBM ranking is fitted once per dataset and seed on the full training pool and held
fixed through each sweep. Points are mean paired error reductions over four seeds, relative to
using all available columns at that point; bars show one standard error. QSAR-TID-11 uses RMSE.}
\label{fig:column-controlled-sweeps}
\end{figure}

\paragraph{Varying rows and columns separately.}
To test whether many columns and few rows explain the benefit, we vary these dimensions
separately (Figure~\ref{fig:column-controlled-sweeps}). The row sweep fixes all $D$ columns and
randomly samples $n\in\{50,100,200,400,800,1600,3200\}$ training rows, up to the available pool
size, on Bioresponse, hiva\_agnostic, CIFAR\_10, and QSAR-TID-11. Each of four seeds defines a
70/15/15 train/unused-validation/test split; classification splits are stratified. Training pools
are capped at 6{,}000 rows. Context sampling is stratified for classification and
uniform without replacement for regression. Both arms use the same sampled rows and fixed
test set: one single view receives all $D$ columns, and the other receives the top 200.
LightGBM ranks columns once on the full training pool, and the selected set stays fixed across
$n$. Thus the sweep changes the rows given to TabPFN without requiring the selector to learn
the ranking from only $n$ labels.

The column sweep fixes the training rows on amazon-reviews, micro-mass, and 11\_Tumors.
Every input retains the same top-200 columns and adds randomly drawn lower-ranked columns,
with total widths from 500 to 10{,}000 where available. A single view using these columns is
compared with a single view using only the fixed top 200. For both sweeps, we compute
$100(1-e_{\rm selected}/e_{\rm full})$ within each seed and then average over seeds.
Small row contexts increase selection gains on Bioresponse, hiva\_agnostic, and QSAR-TID-11,
while CIFAR\_10 shows little change; the trends are not monotonic in $D/n$.
Lower importance rank is not assumed to imply absence of predictive information.

\paragraph{Sampling without class stratification.}
We repeat the row sweep on the three classification tables without balancing class counts.
For each seed, we shuffle the training pool once and take its first $n$ rows, so smaller
contexts are nested within larger ones. The outer splits, test sets, full-pool feature rankings,
200-column budget, and single-view evaluation are unchanged.
For classification, stratified sampling and unstratified nested sampling differ in both class
stratification and overlap across context sizes, so their differences cannot be attributed
solely to stratification.

The four-seed results remain mixed. For example, at $n=100$, Bioresponse's mean error reduction
changes from $9.20\%$ to $-2.66\%$, while CIFAR\_10's unstratified means range from $-0.49\%$
to $1.39\%$ across the tested context sizes. The unstratified sweep does not show consistently
larger benefits from column selection.

On hiva\_agnostic, the unstratified contexts omit at least one test-set class in three of four
seeds at $n=50$, two at $n=100$, and one at $n=200$. We retain these draws and assign zero
predicted probability to unseen classes, using the same float32 log-loss clipping as in the
complete-class cases. Both column settings share the missing-class penalty, so changes in
their relative gains also reflect class coverage.

\subsection{Context expansion}
\label{app:context-generation}

\paragraph{Operator families and coverage.}
\emph{Blind synthesis} adds jittered or interpolated training rows. \emph{Transduction} adds
observed query features with model-generated labels. \emph{Learned rows} optimize synthetic
features through the frozen backbone, with fixed labels except in the joint regression settings.
\emph{Derived columns} append out-of-fold target encodings or backbone predictions.
Table~\ref{tab:expansion-operators} lists 18 expansion settings, an initialization control, and
a curation reference over 19 datasets; Table~\ref{tab:expansion-protocol} defines their construction
and use.

The learned-row summary in Table~\ref{tab:expansion-summary} covers 171 free-row and generator
cases from 51 datasets at $M\in\{32,128,512\}$, with coverage varying by dose.
Full-size joint rows are a separate five-dataset experiment. Row controls, fold-rotated rows,
and compact joint rows appear in Table~\ref{tab:expansion-operators} but are excluded from the
family summary.

The summary counts dataset--operator pairs for blind synthesis and transduction, datasets for
full-size joint rows and derived columns, and probe--dataset--dose cases for learned rows.
For learned rows, 10 of 171 cases meet the decision criterion: mean error reduction above
$0.5\%$ and improvement on more than $60\%$ of evaluated splits. This is an effect-and-consistency
threshold, not a significance test; 54 of 171 cases show any improvement. Other families count
any improvement over the unmodified context. Transduction includes the whole-query pseudo-label control.

\begin{table}[!htb]
\centering
\small
\caption{Context expansion by operator family, frozen TabPFN-3. Relative error reduction against the
unmodified context, in per cent; positive is better, as in Tables~\ref{tab:column-curation}
and~\ref{tab:row-construction}. The scope column names the unit being counted, which differs by
family. \emph{Passes criterion} counts units meeting that family's own bar: an improvement over the
unmodified context for every row except the learned-row program, where it is that program's
decision criterion.}
\label{tab:expansion-summary}
\begin{tabular}{@{}llrrl@{}}
\toprule
\textbf{Family} & \textbf{Scope} & \textbf{Passes} & \textbf{Median} & \textbf{Example/control} \\
\midrule
Blind synthesis      & 51 dataset\,$\times$\,operator & 3  & $-4.6$  & none \\
Transduction         & 57 dataset\,$\times$\,operator & 14 & $-1.8$  & anneal $+12.8$, airfoil $+1.2$ \\
Learned rows         & 171 probe/dataset/dose & 10 & $-0.1$ & concrete improves over placebo \\
Full-size joint rows & 5 datasets & 0 & $-11.8$ & no improvement observed \\
Self-stacking        & 19 datasets & 1  & $-3.0$  & Marketing\_Campaign only \\
OOF target encoding  & 9 datasets  & 2  & $-0.2$  & Marketing\_Campaign $+9.4$ \\
\bottomrule
\end{tabular}
\end{table}

{\small
\setlength{\tabcolsep}{4pt}
\begin{longtable}{@{}>{\raggedright\arraybackslash}p{.30\linewidth}lrrlr@{}}
\caption{Expansion settings and controls over the 19-dataset evaluation. Relative error reduction
against the unmodified context, in per cent; positive is better. $n$ is the number of datasets on
which the operator ran; some operators apply only to specific task types or class structures.
\emph{Improves} counts datasets with any reduction. For learned rows, this differs from the
decision criterion in Table~\ref{tab:expansion-summary}. $M$ is the number of added rows.
Nearest-row curation selects 70\% of the training rows nearest each query-cluster centroid
as a curation reference.}
\label{tab:expansion-operators}\\
\toprule
\textbf{Operator} & \textbf{Family} & $n$ & \textbf{Median} & \textbf{IQR} & \textbf{Improves} \\
\midrule
\endfirsthead
\toprule
\textbf{Operator} & \textbf{Family} & $n$ & \textbf{Median} & \textbf{IQR} & \textbf{Improves} \\
\midrule
\endhead
\bottomrule
\endlastfoot
Gaussian jitter & Blind synthesis    & 19 & $-6.19$ & $[-11.01, -3.99]$ & 2/19 \\
Row interpolation & Blind synthesis    & 19 & $-2.95$ & $[-5.49, -1.84]$ & 1/19 \\
Class-balanced interpolation & Blind synthesis    & 13 & $-5.90$ & $[-20.02, -3.39]$ & 0/13 \\
Whole-query pseudo-labels & Transduction       & 19 & $-12.66$ & $[-44.80, -3.44]$ & 3/19 \\
Cross-half pseudo-labels & Transduction       & 19 & $-2.26$ & $[-2.76, +0.04]$ & 5/19 \\
Cross-half, confidence $\geq0.9$ & Transduction       & 14 & $-0.59$ & $[-1.41, -0.33]$ & 2/14 \\
Cross-half, interval filtering & Transduction       &  5 & $+0.05$ & $[+0.05, +0.09]$ & 4/5 \\
Class duplication with jitter & Row control        & 14 & $-0.01$ & $[-0.13, +0.09]$ & 7/14 \\
Noise rows, class proportions & Row control        & 14 & $-0.04$ & $[-0.41, +0.05]$ & 5/14 \\
Noise rows, majority class & Row control        & 14 & $-0.21$ & $[-0.30, -0.00]$ & 4/14 \\
Target-tail duplication & Row control        &  3 & $+0.04$ & $[-0.17, +0.15]$ & 2/3 \\
Fold-rotated rows & Learned rows       & 19 & $-0.06$ & $[-0.15, +0.01]$ & 6/19 \\
Distilled rows ($M=32$) & Learned rows       & 19 & $-0.10$ & $[-0.19, +0.16]$ & 9/19 \\
Initial rows ($M=32$) & Init.\ control     & 19 & $-0.06$ & $[-0.18, +0.12]$ & 8/19 \\
Generated rows ($M=32$) & Learned rows       & 19 & $-0.02$ & $[-0.27, +0.07]$ & 7/19 \\
Joint rows ($M=32$) & Learned rows       &  5 & $+0.03$ & $[-0.14, +0.10]$ & 3/5 \\
Joint rows (full size) & Full-size joint    &  5 & $-11.76$ & $[-12.97, -8.01]$ & 0/5 \\
OOF target encoding & Derived columns    &  9 & $-0.23$ & $[-0.47, -0.04]$ & 2/9 \\
Self-stacking & Derived columns    & 19 & $-3.00$ & $[-5.19, -0.65]$ & 1/19 \\
Nearest-row curation (70\%) & Curation reference & 19 & $-1.55$ & $[-2.51, -0.47]$ & 3/19 \\
\end{longtable}
}

{\small
\setlength{\tabcolsep}{4pt}
\begin{longtable}{@{}>{\raggedright\arraybackslash}p{.19\linewidth}>{\raggedright\arraybackslash}p{.32\linewidth}>{\raggedright\arraybackslash}p{.23\linewidth}>{\raggedright\arraybackslash}p{.18\linewidth}@{}}
\caption{Construction and use of expansion settings. All backbone weights are frozen.
$\mathcal D$ denotes the outer-training table and $Z$ denotes constructed rows.
Only transductive settings use query features to construct the conditioning data;
query labels are reserved for evaluation.}
\label{tab:expansion-protocol}\\
\toprule
\textbf{Setting} & \textbf{Training / construction} & \textbf{Deployment} & \textbf{Labels / selection} \\
\midrule
\endfirsthead
\toprule
\textbf{Setting} & \textbf{Training / construction} & \textbf{Deployment} & \textbf{Labels / selection} \\
\midrule
\endhead
\bottomrule
\endlastfoot
Blind synthesis & Jitter, pair interpolation, or class-balanced interpolation of training rows;
no optimization. & $\mathcal D+Z$, with at most $n/2$ added rows. & Inherited training labels;
interpolated targets for regression mixup. \\
\addlinespace
Transduction & Predict the query batch from $\mathcal D$. For cross-half expansion, split queries into
two parts. & $\mathcal D$ plus pseudo-labeled rows from the other half; the whole-query control adds
the whole batch. & Model predictions only; optional probability or quantile-width filter. \\
\addlinespace
Row controls & Duplicate training rows or sample noise anchors;
no optimization. & $\mathcal D$ plus the constructed rows. & Training labels or target tails
for duplication; labels for noise anchors sampled from training class proportions or fixed to
the majority class. \\
\addlinespace
Fold-rotated rows & Optimize free rows using real training folds plus $Z$ as context; score a
different training fold. & $\mathcal D+Z$. & Fixed seed-row labels; training-side
early stopping. \\
\addlinespace
Free-row distillation & Optimize row features with $Z$ alone as context; predict real training
queries. & $\mathcal D+Z$; $Z$ alone is a diagnostic arm. & Fixed seed-row labels;
Nominal 10\% training-side checkpoint slice; initialization details below. \\
\addlinespace
MLP-generated rows & Same distillation objective; optimize a conditional MLP with a fixed
latent bank. & $\mathcal D$ plus generated rows. & Fixed conditioning labels from training
rows; training-side checkpoint selection. \\
\addlinespace
Joint rows & Distill with $Z$ alone, optimizing both features and continuous targets;
$M=32$ or the full training size. & $\mathcal D+Z$. & Regression training targets supervise
the objective; same checkpoint rule. \\
\addlinespace
Derived columns & Five-fold training-side target statistics or backbone predictions form new
columns. & Append columns to real context and query features. & Out-of-fold values for training
rows; full-training fit for queries. \\
\end{longtable}
}

\paragraph{Transductive and row-control definitions.}
Prior work also investigates the use of unlabeled examples through repeated application of
tabular foundation models \citep{li2025unlabeled}.
Algorithm~\ref{alg:cross-half} specifies cross-half expansion: its pseudo-labels are fixed predictions
from the original context, and each query receives pseudo-rows only from the other half of the
unlabeled batch. The whole-query control instead includes the whole batch's pseudo-labels in
every query's context. Classification confidence filtering retains pseudo-rows with maximum predicted
class probability at least $0.9$. For regression, interval filtering uses
$(q_{0.9}-q_{0.1})/\operatorname{sd}(y_{\rm train})$ and retains the half with width at or below
the batch median before cross-half prediction. Neither filter reads true query labels.
Gaussian jitter has standard deviation $0.05$ times the training-column standard deviation.
Class-balanced interpolation uses random within-class pairs. The class-duplication control
adds four sampled rows per class with jitter. Noise-row controls add 32 rows labeled by
training class proportions or the majority class. Target-tail duplication adds at most
64 jittered rows from the outer 5\% target tails.
These fixed row controls are separate from the optimized-row program.

\input{algorithms/cross_half_pseudolabeling}

\paragraph{Derived-column construction.}
Self-stacking first predicts each training fold from the other four folds, producing one
probability column per class or one regression-mean column. Predictions from a fit on all
training rows supply the corresponding query columns. A new frozen predictor then uses the
original features concatenated with these columns. Target encoding follows the same five-fold
construction, replacing each categorical value by a smoothed target mean computed on the other
folds: $(n_a\bar y_a+10\bar y)/(n_a+10)$ for category $a$, with the fold's overall mean $\bar y$
for unseen categories. Query encodings use full-training statistics. Thus the new training
features exclude each row's own target; neither construction uses query labels.

\paragraph{Datasets and optimization.}
The 19-\allowbreak dataset matrix covers \texttt{blood-\allowbreak transfusion-\allowbreak service-\allowbreak center}; \texttt{diabetes}; \texttt{credit-\allowbreak g}; \texttt{qsar-\allowbreak biodeg}; \texttt{Fitness\_\allowbreak Club}; \texttt{Is-\allowbreak this-\allowbreak a-\allowbreak good-\allowbreak customer}; \texttt{Marketing\_\allowbreak Campaign}; \texttt{hazelnut-\allowbreak spread-\allowbreak contaminant-\allowbreak detection}; \texttt{seismic-\allowbreak bumps}; \texttt{anneal}; \texttt{maternal\_\allowbreak health\_\allowbreak risk}; \texttt{website\_\allowbreak phishing}; \texttt{MIC}; \texttt{students\_\allowbreak dropout\_\allowbreak and\_\allowbreak academic\_\allowbreak success}; \texttt{QSAR\_\allowbreak fish\_\allowbreak toxicity}; \texttt{concrete\_\allowbreak compressive\_\allowbreak strength}; \texttt{healthcare\_\allowbreak insurance\_\allowbreak expenses}; \texttt{airfoil\_\allowbreak self\_\allowbreak noise}; \texttt{Another-\allowbreak Dataset-\allowbreak on-\allowbreak used-\allowbreak Fiat-\allowbreak 500}.

The broader learned-row program comprises 95 classification, 32 regression, and 44 generator
probe--dataset--dose cases. Free-row distillation uses up to 300 Adam steps at learning rate $3\times10^{-2}$ and
optimizes synthetic inputs as the sole training
context against real training queries, with a training-side checkpoint slice. Deployment appends the
result to the complete real context. This evaluates independently distilled rows as augmentation;
its objective does not optimize the real-plus-synthetic deployment context. The fold-rotated setting includes real context during optimization, rotating over four training
folds with a training-side checkpoint slice. It uses
$M=\operatorname{clip}(\operatorname{round}(0.1N/d),8,N)$ synthetic rows for $N$ training
rows and $d$ features. Generator distillation uses a two-layer, 64-unit MLP and a
fixed latent bank, optimized for up to 400 steps at learning rate $3\times10^{-3}$.
The initialization is evaluated separately as a placebo. Optimization uses two differentiable views;
evaluation uses eight views with constructed rows mapped back to raw feature space. Experiments
use the official outer split grids, with coverage varying by operator, dataset, and row dose.

Algorithm~\ref{alg:free-row-distillation} spells out the free-row classification and regression
probes. The optimization loss $L$ is cross-entropy for classification and mean squared error
of the predictive mean for regression. The checkpoint slice has
$\max(20,\min(2048,\lfloor0.1|D|\rfloor))$ rows. Classification initializes from class-proportional
seed draws over the complete outer-training table, including the checkpoint slice, with at least
one seed per class; rounded quotas can make the realized row count differ from the nominal dose
$M$. Regression draws seeds only from the optimization-query partition. Consequently, the
classification checkpoint slice is excluded from gradient-loss queries but not from seed selection.
Continuous seed features receive Gaussian noise of standard deviation $0.05$ in standardized
units; categorical entries and seed labels remain fixed. The penalty $\Omega$ combines squared
distance to the unperturbed seed rows (coefficient $10^{-3}$) and a soft box penalty outside
$[-3,3]$ (coefficient $10^{-2}$), applied to continuous entries. Checkpoint improvement tolerances
are $10^{-5}$ for classification and $10^{-7}$ for regression. The initialization is itself a
checkpoint candidate. Free-row runs use patience 4 checkpoint checks; the generator uses
patience 5 and an improvement tolerance of $10^{-7}$ for both task types.
Real rows are absent from the optimization context and are restored for
the final augmentation evaluation. The fold-rotated and generator settings retain the distinct
objectives and parameterizations in Table~\ref{tab:expansion-protocol}.

\input{algorithms/free_row_distillation}

\paragraph{The placebo control.}
Optimised rows change two things at once, since adding $M$ rows perturbs the context whether or not
those rows were optimised. Every learned-row cell is therefore paired with a matched arm that adds
the same number of rows at their initialisation, before optimisation. The difference between the two
is the part attributable to the optimisation; the initialised arm is the part attributable to having
more rows at all. For free-row distillation on concrete at $M=32$, the optimized context lowers error by
$0.684\%$, whereas its initialization raises it by $0.063\%$: a $0.747$ percentage-point benefit
from optimization. In some other cases, improvements also occur with the unoptimized initialization.

\paragraph{Results by dose.}
Coverage differs across doses, so these are separate descriptive summaries. At $M=32$, the
median reduction is $-0.03\%$: 28 of 72 cases improve, and 7 meet the decision criterion.
At $M=128$, the median is $-0.13\%$: 21 of 69 improve, and 2 meet the criterion.
At $M=512$, the median is $-0.12\%$: 5 of 30 improve, and 1 meets the criterion.

%% file: algorithms/model_aware_retrieval.tex
\begin{algorithm}[!t]
\caption{Model-aware row retrieval with optional sharded scoring}
\label{alg:model-aware-retrieval}
\small
\begin{algorithmic}[1]
\Require Labeled pool $D=\{(\vx_i,y_i)\}_{i=1}^N$, unlabeled query batch $X_Q$, frozen $f_\theta$
\Require Retrieval budget $r$, query-cluster count $G$, block size $B$, union cap $C$, deployment views $V$
\Require Overflow policy $\pi\in\{\mathrm{split},\mathrm{truncate}\}$ (split for Higgs)
\Ensure Predictions for $X_Q$ in its original order
\State Partition the shuffled pool indices into blocks $B_1,\ldots,B_J$ of at most $B$ rows
\State $r\gets\min(r,N)$; $u_b\gets r|B_b|/N$; $r_b\gets\lfloor u_b\rfloor$ for each block $b$
\State Add one to the $r-\sum_b r_b$ quotas with largest fractional remainder $u_b-r_b$
\For{each block $b$ with $r_b>0$}
  \State Fit a frozen single-view predictor on $D_{B_b}$ and process $X_Q$
  \State Stream last-ICL-layer query-to-context softmax attention, averaged over heads
  \State For each query $j$, retain the top-$r_b$ pool indices $S_{jb}$ and their attention values
\EndFor
\State $S_j\gets\bigcup_b S_{jb}$ for each query $j$; retain the block-local attention values $a_{ji}$
\State Form sparse binary indicators $H_{ji}=\mathbf{1}\{i\in S_j\}$ over the retrieved pool indices
\State $(Q_1,\ldots,Q_G)\gets\operatorname{KMeans}(H,G)$
\State Initialize a queue with the nonempty query clusters
\While{the queue is nonempty}
  \State Pop a query cluster $Q_g$
  \State $U_g\gets\bigcup_{j\in Q_g}S_j$
  \If{$|U_g|>C$ and $\pi=\mathrm{split}$}
    \State Fit two KMeans centers to $H_{Q_g}$; order queries by their distance difference
    \State Push the two balanced halves of this ordering onto the queue
  \Else
    \If{$|U_g|>C$}
      \State $s_i\gets\sum_{j\in Q_g:\,i\in S_j}a_{ji}$ for $i\in U_g$
      \State Retain in $U_g$ only the $C$ rows with largest $s_i$
    \EndIf
    \State Fit a frozen $V$-view predictor on $D_{U_g}$ and predict $X_{Q_g}$
    \State Align classification probabilities to the pool's class order and insert into query order
  \EndIf
\EndWhile
\State \Return the assembled predictions
\end{algorithmic}
\end{algorithm}

%% file: algorithms/cross_half_pseudolabeling.tex
\begin{algorithm}[!htbp]
\caption{Cross-half pseudo-label expansion}
\label{alg:cross-half}
\small
\begin{algorithmic}[1]
\Require Labeled training table $D$, unlabeled query batch $X_Q$, frozen predictor $f_\theta$
\Require Optional confidence rule and a fixed random seed
\Ensure Predictions for every query, without using its own pseudo-label as context
\State $P^0\gets\operatorname{Predict}_{\theta,8}(X_Q;D)$
\State Form fixed pseudo-labels $\tilde y_j$ from $P^0$: most probable class or regression mean
\State Mark eligible pseudo-rows using the confidence rule, or retain all if none is specified
\State Randomly partition query indices into halves $Q_1,Q_2$
\For{$h=1,2$}
  \State $Z_h\gets\{(\vx_j,\tilde y_j):j\in Q_{3-h},\ j\text{ is eligible}\}$
  \State $\hat P[Q_h]\gets\operatorname{Predict}_{\theta,8}(X_{Q_h};D\cup Z_h)$
\EndFor
\State \Return $\hat P$ in original query order
\end{algorithmic}
\end{algorithm}

%% file: algorithms/free_row_distillation.tex
\begin{algorithm}[!htbp]
\caption{Free-row distillation evaluated as context augmentation}
\label{alg:free-row-distillation}
\small
\begin{algorithmic}[1]
\Require Outer-training table $D$, unlabeled queries $X_Q$, frozen $f_\theta$, row dose $M$
\Ensure Augmented-context predictions and matched initialization-control predictions
\State Standardize features using $D$; split its rows into optimization queries $D_o$ and checkpoint slice $D_h$
\State Initialize synthetic rows $Z^0=(X_Z^0,y_Z)$ at nominal dose $M$, as specified in the text
\State $Z\gets Z^0$; freeze $\theta$, $y_Z$, and categorical entries; retain $Z^0$ for the initialization control
\State $Z^\star\gets Z^0$; evaluate its checkpoint loss using $Z^0$ alone as context
\For{at most 300 optimization steps}
  \State Draw a query minibatch $(X_b,y_b)$ from $D_o$ (use all if there are at most 4096 rows)
  \State $\mathcal L\gets L(f_\theta(X_b;Z),y_b)+\Omega(X_Z)$, using two differentiable views
  \State Take an Adam step on the continuous entries of $X_Z$, with learning rate $3\times10^{-2}$
  \If{this is a checkpoint step (every 25 steps)}
    \State Evaluate $L(f_\theta(X_h;Z),y_h)$ on $D_h$, using $Z$ alone as context
    \State Save a copy of $Z$ as $Z^\star$ if its loss improves; stop after four unsuccessful checks
  \EndIf
\EndFor
\State Map $Z^\star$ and $Z^0$ back to raw feature space
\State \Return $\operatorname{Predict}_{\theta,8}(X_Q;D\cup Z^\star)$ and $\operatorname{Predict}_{\theta,8}(X_Q;D\cup Z^0)$
\end{algorithmic}
\end{algorithm}

%% file: appendix_cross_analysis.tex
\section{Additional cross-intervention analyses}
\label{app:cross-analysis}

\subsection{Composition of interventions}
\label{app:composition}

\paragraph{Method and information flow.}
The composed arm applies 96-configuration greedy aggregation after DiagScale adaptation of
TabPFN-3. One adapter is fitted per outer cell at learning rate $10^{-2}$ for up to 30 epochs,
using an internal 90/10 split for fitting and checkpoint selection. The pretrained weights remain
frozen. Algorithm~\ref{alg:aggregation} uses this shared pretrained-plus-adapter state as
$\bar\theta$ for every configuration during both inner-fold prediction and testing. Inner folds
rebuild preprocessing and contexts without retraining the adapter.


\paragraph{Interaction on log error.}
Elo is a fitted, field-dependent scale, so adding Elo gains does not establish additivity in
predictive performance. We measure interaction on log error, comparing four independently
executed recipes on the same 816 outer-test cells:
the frozen backbone, DiagScale, the frozen 96-configuration pool, and DiagScale plus that pool.
Their descriptive interaction on log error is
\begin{equation}
\label{eq:interaction}
I \;=\; \big[\log e(\text{pool}+\text{DiagScale}) - \log e(\text{DiagScale})\big] \;-\;
        \big[\log e(\text{pool}) - \log e(\text{base})\big],
\end{equation}
where each log error is first averaged over cells within a dataset, then over the 51 equally
weighted datasets. Positive $I$ indicates less reduction than the sum of the standalone gains.
Intervals use 10{,}000 dataset bootstrap samples, conditional on these selected recipes.
The standalone adapter and composition use separate fitting and inference paths; the comparison
does not assume that the composition's configuration~0 reproduces the standalone predictions
exactly.

\paragraph{Result.} DiagScale lowers mean log error by $0.007420$, the pool by $0.024774$,
and the composition by $0.031226$. The interaction is
$+0.000968$ $[-0.002323,+0.003923]$; this interval does not establish equivalence to zero.
The composition's error change is $-2.352\%$ $[-4.557\%,-0.742\%]$ relative to DiagScale and
$-0.643\%$ $[-0.973\%,-0.353\%]$ relative to aggregation, using the exponentiated mean log
ratio. It improves over both on 27 of 51 datasets.

\paragraph{Sensitivity to the default member.}
We also score configuration~0 from both pools on all 816 cells. Relative to the standalone
frozen and DiagScale runs, its error changes are $-0.038\%$ $[-0.116\%,+0.002\%]$ and
$-0.076\%$ $[-0.353\%,+0.149\%]$, respectively. Using these pool-internal references gives an
interaction of $+0.001340$ $[-0.000037,+0.003170]$, and the composition still lowers error by
$2.279\%$ $[0.704\%,4.460\%]$ relative to its own adapted default member. This sensitivity
check uses each pool's own reference predictions.

\subsection{Concentration of gains}
\label{app:gain-concentration}
For each dataset, we compute $100(1-\bar e_{\rm arm}/\bar e_{\rm base})$, using arithmetic mean
cell errors in both numerator and denominator. The top-five share divides the sum of the five
largest positive reductions by the sum of all positive reductions. The inputs are the seven
deployed outcomes, including validation-selected fallback for adaptation.
This ratio-of-means estimator yields 41 aggregation wins; the mean-log-ratio estimator in
Table~\ref{tab:agg-canonical} yields 42. Table~\ref{tab:concentration} summarizes gain concentration.

\begin{table}[h]
\centering\small
\caption{Concentration of positive per-dataset error reductions.}
\label{tab:concentration}
\begin{tabular}{@{}llrr@{}}
\toprule
Backbone & Method & Top-5 share & Datasets improved\\
\midrule
TabFM & DiagScale & $70\%$ & 25/51 \\
TabFM & full fine-tuning & $62\%$ & 31/51 \\
TabICL v2 & DiagScale & $46\%$ & 34/51 \\
TabICL v2 & full fine-tuning & $48\%$ & 35/51 \\
TabPFN-3 & DiagScale & $56\%$ & 31/51 \\
TabPFN-3 & aggregation & $68\%$ & 41/51 \\
TabPFN-3 & full fine-tuning & $59\%$ & 37/51 \\
\bottomrule
\end{tabular}
\end{table}

\subsection{Predicting gains}
\label{app:predicting-gains}
\paragraph{Design.}
We screen 23 predictor columns representing 21 distinct rank tests against seven outcomes
(147 tests): DiagScale and full fine-tuning on three backbones, and 96-configuration greedy
aggregation on TabPFN-3. The logarithms of training size and feature count duplicate their
Spearman ranks and are counted only once. Predictors include dataset size, width, task type,
class balance, missingness, cardinality, feature correlation, univariate signal concentration,
frozen-model error and calibration, and view disagreement. Outcomes use the ratio-of-means
estimator above. This screen and the task-type sensitivity analysis are exploratory.

All behavior-based covariates use TabPFN-3 predictions, including when predicting TabICL or TabFM
adaptation. Classification disagreement is mean pairwise Jensen--Shannon divergence (base two)
among native-view probabilities, averaged over query rows. Regression disagreement is mean
across-view prediction variance normalized by the variance of training-side labels. Frozen error,
its fold variation, and calibration use training-side out-of-fold predictions; feature statistics
use training rows. Query labels do not enter the predictors. Error and calibration have different
interpretations across task types, motivating the stratified sensitivity analysis below.

\paragraph{Univariate screen.}
We report Spearman correlations and Benjamini--Hochberg adjusted values both within each outcome
(21 tests) and globally (147 tests). At $q\leq0.10$, native-view disagreement is the only predictor
passing a within-outcome screen: for aggregation and both TabICL adaptations. Under global BH,
only its aggregation association passes. Table~\ref{tab:predictor-canonical} gives all seven
associations.

\begin{table}[h]
\centering\small
\caption{Native-view disagreement and intervention gains. Disagreement uses TabPFN-3 predictions
for all outcomes. $\rho_{32}$ is its Spearman correlation with per-dataset error reduction using
32 views; $q_{21}$ and $q_{147}$ are Benjamini--Hochberg adjusted $p$-values within each outcome
and across all outcomes, respectively. $q_{21,8}$ repeats the within-outcome correction using
eight views, four from each of the two preprocessing pipelines.}
\label{tab:predictor-canonical}
\begin{tabular}{@{}lrrrr@{}}
\toprule
Outcome & $\rho_{32}$ & $q_{21}$ & $q_{147}$ & $q_{21,8}$\\
\midrule
TabFM / DiagScale & 0.212 & 0.282 & 0.599 & 0.358 \\
TabFM / full FT & 0.284 & 0.906 & 0.528 & 0.941 \\
TabICL v2 / DiagScale & 0.426 & 0.039 & 0.130 & 0.123 \\
TabICL v2 / full FT & 0.412 & 0.056 & 0.130 & 0.138 \\
TabPFN-3 / DiagScale & 0.339 & 0.317 & 0.444 & 0.264 \\
TabPFN-3 / aggregation & 0.496 & 0.004 & 0.031 & 0.013 \\
TabPFN-3 / full FT & 0.277 & 0.581 & 0.528 & 0.581 \\
\bottomrule
\end{tabular}
\end{table}

\paragraph{Measurement budget and task type.}
The balanced eight-view variant has an aggregation association of $\rho=0.462$, with
within-outcome $q=0.013$. The 32-view result is $\rho=0.496$, with global $q=0.031$.
The association varies across task types: for aggregation it is $+0.567$ on 30 binary tasks,
$+0.231$ on 13 regression tasks, and $-0.310$ on eight multiclass tasks. These post-hoc subgroup
estimates are descriptive; the small groups and differing disagreement definitions limit their
interpretation.

\paragraph{Multivariate prediction.}
We repeat five-fold cross-validation over datasets 20 times with fixed seeds, using gradient-boosted
trees and ridge regression, with rank transformation applied only to ridge. Imputation and
transformations are fitted within training folds. Among 14 model--outcome combinations, 13 have negative mean out-of-fold
$R^2$. The exception is ridge for aggregation, $R^2=0.012$ and mean Spearman $\rho=0.299$;
boosted trees give $R^2=-0.083$ on that outcome. Accuracy in predicting gain magnitudes
remains limited despite the univariate association.

\subsection{Runtime and cost accounting}
\label{app:runtime-comparison}
\label{app:runtime-accounting}

\begin{table}[!htb]
\centering\small
\caption{Measured runtime over the same 816 TabArena cells on single H100 GPUs. Fit includes
adaptation or out-of-fold member evaluation and selection; predict covers the full test batches.
Pool-based methods include predictions from all 96 configurations. Elo uses 68 fixed references
and one candidate per fit; $\Delta$ Elo uses the corresponding baseline within that fit.
Frozen TabPFN-3 and TabICL v2 compare reproduced models with published defaults, so their deltas
need not be zero. Frozen TabFM uses its rating in the TabFM full-FT fit.
Runtime multiples use the reproduced frozen models.}
\label{tab:cost}
\begin{tabular}{@{}lrrrrrr@{}}
\toprule
\textbf{Recipe} & \textbf{Elo} & \textbf{$\Delta$ Elo} & \textbf{Fit (h)} & \textbf{Predict (h)} & \textbf{Total (h)} & \textbf{$\times$ frozen} \\
\midrule
\multicolumn{7}{@{}l}{\emph{TabPFN-3}} \\
Frozen & 1658 & $+1$ & $0.14$ & $0.38$ & $0.52$ & $1.0$ \\
32 native views & 1675 & $+21$ & $0.19$ & $1.47$ & $1.66$ & $3.2$ \\
DiagScale & 1682 & $+26$ & $6.30$ & $0.31$ & $6.61$ & $12.7$ \\
Full fine-tuning & 1681 & $+25$ & $7.42$ & $0.31$ & $7.73$ & $14.9$ \\
Aggregation (96) & 1721 & $+67$ & $177.71$ & $69.79$ & $247.50$ & $476.4$ \\
DiagScale + aggregation & 1748 & $+93$ & $184.02$ & $69.07$ & $253.09$ & $487.2$ \\
\midrule
\multicolumn{7}{@{}l}{\emph{TabICL v2}} \\
Frozen & 1592 & $+9$ & $0.15$ & $1.08$ & $1.23$ & $1.0$ \\
DiagScale & 1664 & $+82$ & $9.68$ & $1.08$ & $10.76$ & $8.7$ \\
Full fine-tuning & 1659 & $+77$ & $9.85$ & $1.04$ & $10.88$ & $8.8$ \\
\midrule
\multicolumn{7}{@{}l}{\emph{TabFM}} \\
Frozen & 1792 & $+0$ & $0.12$ & $19.19$ & $19.32$ & $1.0$ \\
DiagScale & 1812 & $+20$ & $11.23$ & $48.14$ & $59.37$ & $3.1$ \\
Full fine-tuning & 1820 & $+28$ & $13.89$ & $48.00$ & $61.89$ & $3.2$ \\
\bottomrule
\end{tabular}
\end{table}

Figure~\ref{fig:cost-performance} uses the totals in Table~\ref{tab:cost}; phase shares divide
each phase's summed time by the summed total.

For configuration aggregation, out-of-fold member predictions account for 177.6 hours,
approximately 72\% of the 247.5-hour total. The total includes test predictions from all 96 configurations;
selected-member serving was not timed. The label-fraction study uses those fixed
predictions and does not measure runtime.